\documentclass[10pt,twocolumn,letterpaper]{article}

\usepackage{multirow}
\usepackage[pagenumbers]{cvpr} 

\usepackage{makecell}
\usepackage{xcolor}
\usepackage{url}

\definecolor{cvprblue}{rgb}{0.21,0.49,0.74}
\usepackage[pagebackref,breaklinks,colorlinks,allcolors=cvprblue]{hyperref}

\def\confName{CVPR}
\def\confYear{2026}

\title{CustomTex: High-fidelity Indoor Scene Texturing via Multi-Reference Customization}

\author{
Weilin Chen,
Jiahao Rao,
Wenhao Wang,
Xinyang Li,
Xuan Cheng\thanks{Corresponding author: Xuan Cheng (chengxuan@xmu.edu.cn)},
Liujuan Cao \\
Key Laboratory of Multimedia Trusted Perception and Efficient Computing, \\
Ministry of Education of China, Xiamen University\\
{\color{red}\url{https://chenweilinx.github.io/CustomTex/}}
}

\begin{document}
\maketitle
\begin{abstract}
The creation of high-fidelity, customizable 3D indoor scene textures remains a significant challenge. While text-driven methods offer flexibility, they lack the precision for fine-grained, instance-level control, and often produce textures with insufficient quality, artifacts, and baked-in shading. To overcome these limitations, we introduce CustomTex, a novel framework for instance-level, high-fidelity scene texturing driven by reference images. CustomTex takes an untextured 3D scene and a set of reference images specifying the desired appearance for each object instance, and generates a unified, high-resolution texture map. The core of our method is a dual-distillation approach that separates semantic control from pixel-level enhancement. We employ semantic-level distillation, equipped with an instance cross-attention, to ensure semantic plausibility and ``reference-instance'' alignment, and pixel-level distillation to enforce high visual fidelity. Both are unified within a Variational Score Distillation (VSD) optimization framework. Experiments demonstrate that CustomTex achieves precise instance-level consistency with reference images and produces textures with superior sharpness, reduced artifacts, and minimal baked-in shading compared to state-of-the-art methods. Our work establishes a more direct and user-friendly path to high-quality, customizable 3D scene appearance editing.
\end{abstract}    
\vspace{-10pt}
\section{Introduction}
\label{sec:intro}

The creation of photorealistic and stylistically consistent 3D indoor scenes is a cornerstone of applications in virtual and augmented reality, architectural visualization, and film production. A critical factor in achieving this realism is texturing, which entails assigning surface materials and colors to 3D geometry. While recent advancements in 3D reconstruction, such as Neural Radiance Fields \cite{NeRF2020} and 3D Gaussian Splatting \cite{3DGSTOG2023}, have made remarkable progress in capturing geometry and view-dependent appearance, they often produce ``baked-in'' textures that are entangled with lighting and lack material properties. Consequently, re-texturing these scenes with new, consistent and high-quality materials remains a significant challenge.

Most recent 3D indoor scene texturing methods \cite{SceneTex2024, scene2, InstanceTex2024} address this problem by leveraging the power of pre-trained text-to-image diffusion models \cite{diffusion6, StableDiffusion2022}. By using text prompts as guidance, these methods can, in principle, generate a vast array of materials and styles.
However, this approach is often insufficient for indoor scene texturing, as text prompts alone are inherently ambiguous and struggle to convey precise visual characteristics. Using a reference image as prompt \cite{FlexiTex2025} can provide more direct visual guidance, such as the photographs of different home decoration styles. Yet, this approach typically offers only global, coarse-level control, still falling short of allowing users to specify fine-grained attributes like the specific weave of a fabric, the exact grain of wood, or the subtle pattern of a wallpaper.

Moreover, state-of-the-art 3D indoor scene texturing methods \cite{SceneTex2024, scene2, InstanceTex2024} often produce textures of insufficient quality for high-fidelity rendering.
Their outputs lack the sharpness and richness found in artist-created or physically scanned textures, appearing instead soft, blurry or unnaturally uniform upon close inspection. This limitation arises from the entanglement of pixel-level fidelity and semantic-level perception in the diffusion process \cite{PiSA-SR2025}. Additionally, these methods tend to produce textures with ``baked-in'' shading, as diffusion models learn and replicate the lighting and depth cues prevalent in the training datasets. The resulting textures that contain obvious highlights and shadows are not suitable for differently lighted renderings.

To overcome these limitations, we present \textbf{CustomTex}, a novel framework designed for instance-level controllable and high-fidelity texturing of 3D indoor scenes. As shown in Fig. \ref{fig:teaser}, CustomTex generates a customized texture for a 3D scene mesh, based on a set of reference images that specify the desired appearance for each instance (e.g., sofa, cabinet, chair and walls) in the scene.
The texture generated by CustomTex maintains instance-level consistency with the reference images, and yields a visually compelling appearance with significantly reduced blurriness, artifacts, and ``baked-in'' shading compared to existing SOTA methods. To this end, CustomTex separates semantic generation and pixel enhancement into two distinct distillation processes with two pre-trained stable diffusion models \cite{depthad2, PiSA-SR2025}.
The semantic-level distillation, equipped with an instance cross-attention mechanism, ensures sufficient semantic plausibility and enables precise instance-specific control over the generated textures. The pixel-level distillation focuses on boosting the visual fidelity and quality of the generated texture, while preserving the underlying structural and semantic information. Both processes are unified within a single optimization framework based on Variational Score Distillation (VSD) \cite{object3}.

The technical contributions of this paper are summarized as follows:
\begin{itemize}
    \item  A novel texture generation 
    framework that enables flexible, instance-level customization using multiple reference images as prompts.
    
    \item A dual-distillation training approach that effectively preserves semantic content while enhancing pixel-level fidelity.
    
    \item An instance-guided Variational Score Distillation approach that captures the multiple modes present in the reference imagery.
\end{itemize}

\section{Related Work}
\subsection{Neural 3D Generation}
Early research on 3D generation neural methods primarily relied on curated 3D datasets and explicit geometric representations, including voxels~\cite{voxel1, voxel2, voxel3, voxel4, voxel5}, point clouds~\cite{pointcloud1,pointcloud2,pointcloud3, PointsL0}, meshes~\cite{mesh1,mesh2,mesh3,DCT2024,SizeScene2025}, and signed distance fields~\cite{sdf1,sdf2,sdf3,sdf4,sdf5}, which established foundational pipelines for learning 3D structures. However, the scarcity and high cost of high-quality 3D datasets limit current generative models, hindering their scalability, shape diversity, and visual fidelity due to limited scale and category coverage.

Driven by the progress in large-scale vision-language models~\cite{vlm1,vlm2,vlm3,vlm4,vlm5}, recent methods treat 3D as a rendering target rather than a domain requiring explicit 3D supervision. By leveraging differentiable rendering and volumetric neural representations, these methods can propagate gradients directly from 2D observations~\cite{2DI1,2DI2,2DI3,2DI4,Learn2Talk2025}, expanding the effective data regime to infer 3D structure from vast collections of unlabeled images~\cite{2DI6}. However, these methods still faces challenges in reconstructing fine-grained geometry and maintaining high-frequency appearance details.

\subsection{Feed-Forward Texturing}
Texture synthesis aims to produce coherent and high-fidelity appearances on 3D surfaces, ensuring consistency in color~\cite{color1, FaceRefiner2024} and geometry cues~\cite{geometry1,geometry2,geometry3} that reflect shading, material response, and view-dependent effects. 

Unlike 2D texture transfer~\cite{2dt1,2dt2}, 3D texturing must maintain cross-view and geometric consistency over complex surfaces. With the advancement of learning-based 3D representations, UV-based networks~\cite{uv, DNPM}, convolution-based operations~\cite{c1,c2}, transformer-based methods~\cite{transformer1,transformer2} and StyleGAN-inspired architectures~\cite{stylegan, EMEF2023} have jointly modeled geometry and appearance, achieving more faithful texturing for 3D objects and small-scale scenes. Nevertheless, these methods often rely on carefully curated textured datasets and generalize poorly to complex real-world settings, which exhibit diverse geometries, cluttered compositions, and intricate light–material interactions.


\subsection{Diffusion-based Texturing}
Diffusion models~\cite{diffusion6, StableDiffusion2022, DiffSFSR2024,zhang2024fast,zhang2025storyweaver} have revolutionized generative modeling, providing powerful 2D priors that enable high-quality 3D texture generation for both objects~\cite{object1,object2,object3} and scenes~\cite{scene1,scene2} without large-scale 3D supervision, building upon their success in realistic image synthesis.

Existing diffusion-based 3D texturing frameworks can be broadly categorized into three major streams. 
(1) Inpainting-based methods, such as TEXTure~\cite{TEXTure}, Text2Tex~\cite{Text2Tex} and TexFusion~\cite{TexFusion},  progressively generate or refine textures from each rendered view with depth-aware diffusion models~\cite{depthad1,depthad2}. Then, they project and blend the results onto the 3D surface to form a global texture map progressively. While effective, this sequential process often introduces artifacts such as cross-view inconsistencies, seams, and texture drift. 
(2) Texture-space sampling methods like GenesisTex~\cite{genesistex} and GenesisTex2~\cite{genesistex2}, address the coherence issue by performing diffusion directly on UV maps or latent textures, utilizing cross-view latent buffers and multiview score feedback. This strategy yields superior global consistency but introduces a dependence on high-quality UV parameterization and faces scalability issues with large or irregular assets. 
(3) Score Distillation Sampling (SDS) based methods enable direct 3D optimization through gradients from pre-trained diffusion models. Originally introduced by DreamFusion~\cite{dreamfusion} for text-driven NeRF optimization, SDS was later extended to latent-space training in Latent-NeRF~\cite{latentnerf} and texture optimization in Latent-Paint. Subsequent works~\cite{sds1,sds2,sds3} incorporated geometry-aware constraints, enhanced sampling strategies, and accelerated convergence, evolving SDS from slow per-scene optimization toward more scalable and generalizable 3D generation frameworks.

\subsection{3D Indoor Scene Texturing}

SceneTex~\cite{SceneTex2024} is a representative work in this field, which marks an important step toward realistic indoor environment texturing using VSD. RoomPainter~\cite{scene2} features a zero-shot technique that effectively adapts diffusion model for 3D-consistent texture synthesis. However, both methods rely on text prompts, which can't convey precise visual information, and offer only global, coarse control over the texture generation process.
The only existing method for instance-level control is InstanceTex~\cite{InstanceTex2024}, which uses multiple text prompts to specify textures for different objects. Unfortunately, it shares the limitations of other text-driven methods and struggles to produce textures of sufficient quality.


Our method explores image-driven, instance-level texturing for indoor scenes. This paradigm provides fine-grained supervision, material-level alignment and richer style controllability, creating a more direct and user-friendly path to high-fidelity scene appearance editing.

\begin{figure*}
    \centering
    \includegraphics[width=0.95\textwidth]{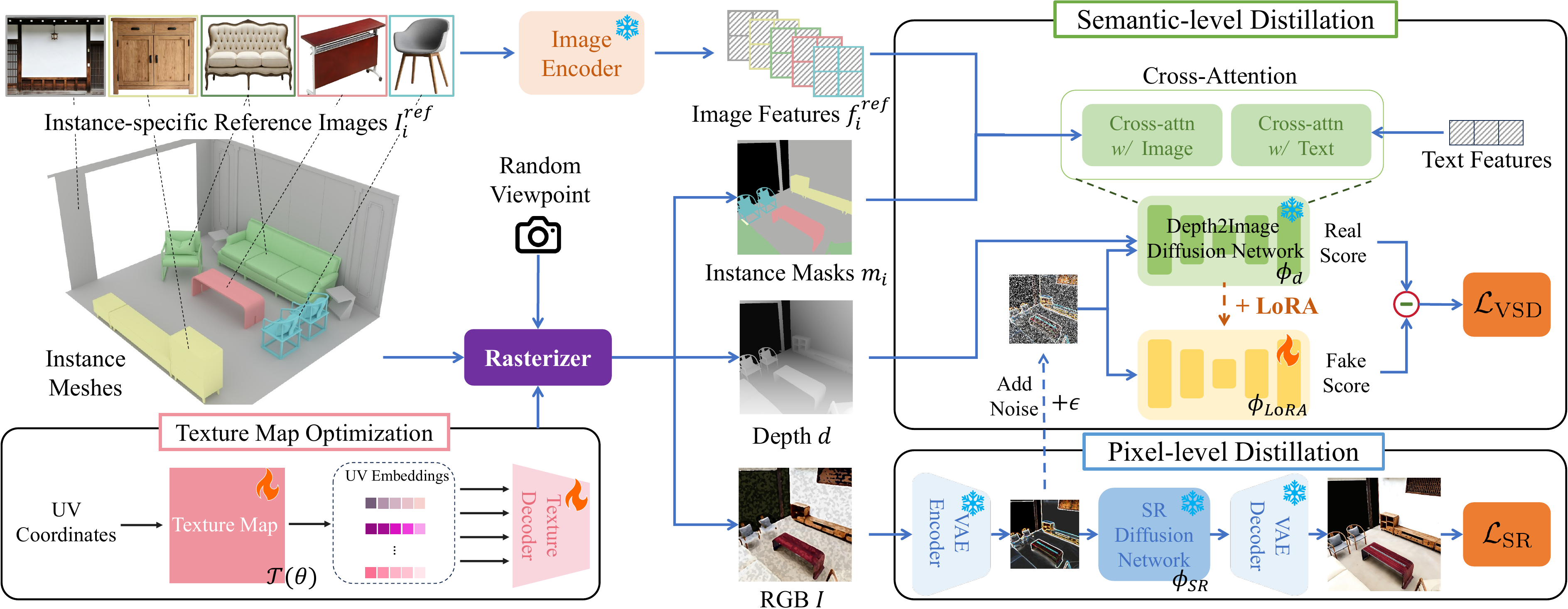}
    \vspace{-5pt}
    \caption{\textbf{Pipeline of CustomTex.} CustomTex textures a complete 3D indoor scene by optimizing a texture map in UV space through a dual-distillation training approach. In each iteration, the 3D scene with optimized texture is rendered from a random viewpoint, producing an RGB image, a depth map and instance masks. Instance masks are used to align each reference image's features with the correct object instance in the rendered RGB image via a specialized cross-attention. The Variational Score Distillation gradient and the Super-Resolution gradient are computed based on the well-aligned reference images condition to update the texture field.}
    \label{fig:pipeline}
    \vspace{-10pt}
\end{figure*}

\section{Method}

\subsection{Overview}
Our work aims to texture a complete 3D indoor scene from a collection of reference images. The framework takes two inputs: an untextured 3D scene composed of multiple object instances $\{\mathcal{O}_i\}_{i=1}^N$, and a set of reference images $\{\mathcal{I}_i^{ref}\}_{i=1}^N$ that define the desired appearance of each instance. We assume that the input 3D scene is unwrapped, where UV coordinates map each vertex of the mesh to a texel in a texture map.
The main requirements for the output texture $\mathcal{T}$ are twofold: 1) each instance's texture must faithfully harmonize with the appearance of its assigned reference image; 2) the entire texture must be of high quality and form a stylistically coherent whole across all instances.

We formulate this texture synthesis task as an optimization problem in the UV space, and propose a dual-distillation training approach comprising semantic-level and pixel-level distillation. As shown in Fig. \ref{fig:pipeline}, the framework distills a pre-trained depth-to-image diffusion model \cite{depthad2} to provide a prior on semantic plausibility based on the reference images prompts. Concurrently, the framework also distills a pre-trained super-resolution diffusion model \cite{PiSA-SR2025} to provide a prior on visual fidelity and quality of the rendered texture. The optimized texture is represented by an implicit multi-resolution texture field that encodes the texture features at different scales in the UV space.

\subsection{Semantic-level Distillation}
To ensure the generated texture for each instance semantically aligns with its reference image, we employ a semantic-level distillation process using instance-guided Variational Score Distillation (InsVSD). InsVSD conditions the distillation on instance-level images prompts, thus allowing the generated texture to learn a comprehensive distribution that encapsulates the multiple modes present in the reference imagery.

The 3D target mesh, with its optimized texture $\mathcal{T}$, is projected to a randomly sampled viewpoint via a differentiable rasterizer, rendering an RGB image $I$, a depth image $d$ and a set of instance masks $\{m_i\}_{i=1}^N$ for different object instances. The depth image $d$ is fed into the depth-to-image diffusion model \cite{depthad2}, which is conditioned on the features $\{f_i^{ref}\}_{i=1}^N$ of reference images $\{\mathcal{I}_i^{ref}\}_{i=1}^N$ to provide semantic guidance.  These features are extracted from the reference images by the image encoder in IP-Adapter \cite{IPAdapter}. To align each feature $f_{i}^{ref}$ with its corresponding instance's position on the rendered RGB image $I$, we use the instance mask $m_{i}$ to adjust the computation of cross attention corresponding to $f_{i}^{ref}$. 
This process is formulated as:
\begin{equation}
\label{eq:masked_cross}
    Z' = \frac{1}{N} \sum_{i=1}^N m_i \cdot \mathrm{Softmax}(\frac{\mathbf{QK}_i^\top}{\sqrt{d_k}})\mathbf{V}_i,
\end{equation}
where $\mathbf{Q}=Z\mathbf{W}_q, \mathbf{K} = f_{i}^{ref}\mathbf{W}_k, \mathbf{V} = f_{i}^{ref}\mathbf{W}_v$ represent the queries, keys and values within the cross-attention module, $Z$, $Z'$ denote the input and output features of the module, and $\mathbf{W}_q, \mathbf{W}_k, \mathbf{W}_v$ are the projection matrices used for linear transformations. As the mask $m_{i}$ can accurately align each $f_{i}^{ref}$ with specific positions of the rendered RGB image $I$, the appearance of the original reference image $I_{i}^{ref}$ is precisely positioned on the relevant instance's parts in the texture UV map $\mathcal{T}$.

In the distillation, we transfer the semantic and stylistic content of each $I_{i}^{ref}$ into the corresponding parts in $\mathcal{T}$. To achieve this, we train a lightweight and learnable LoRA model $\phi_{LoRA}$ to learn the prior from the frozen depth-to-image model $\phi_{d}$. 
This prior defines the characteristics of a plausible rendered RGB image based on the provided image prompts $I_{i}^{ref}$.
As both the parameters $\phi$ in the LoRA model $\phi_{LoRA}$ and the parameters $\theta$ in texture $\mathcal{T}(\theta)$ need to be optimized, we adopt an alternative two-step optimization strategy. Firstly, the LoRA model is frozen and $\theta$ is optimized via the VSD gradient:
\begin{equation}
\label{eq:vsd}
\begin{split}
\nabla_{\theta} \mathcal{L}_{\mathrm{VSD}}(\theta, d, c^{ref}) =\mathbb{E}_{t, \epsilon}[w(t)(
\epsilon_{\phi_d}(\mathcal{T}(\theta); d, c^{ref}, t)- \\
\epsilon_{\phi_{LoRA}}(\mathcal{T}(\theta); d, c^{ref}, t)) 
\frac{\partial \mathcal{T}(\theta)}{\partial \theta}],    
\end{split}
\end{equation}
where $c^{ref}$ denotes the reference image prompts condition, $t$ denotes the time step and $w(t)$ denotes the weighting function. After $\mathcal{T}(\theta)$ is updated via the VSD gradient, we unfreeze LoRA model and update $\phi$ with $\theta$ fixed. The training objective for the LoRA model is defined as:  
\begin{equation}
\label{eq:LoRA}
\begin{split}
\mathcal{L}_{\mathrm{LoRA}}(\phi, d, c^{ref}) &= \\
\min_{\phi} \mathbb{E}_{t, \epsilon}[&||\epsilon_{\phi_{LoRA}}(\mathcal{T}(\theta); d, c^{ref}, t)-\epsilon||_2^2].
\end{split}
\end{equation}
where the added noise is $\epsilon \sim N(0,1)$.


\subsection{Pixel-level Distillation}
While the semantic guidance ensures the instance-level consistency with the reference images, achieving high-fidelity texture details is paramount for visual realism. To this end, our pixel-level distillation leverages a pre-trained image super-resolution model \cite{PiSA-SR2025} to enhance the clarity and resolution of the synthesized texture.
This model \cite{PiSA-SR2025} achieves state-of-the-art super-resolution performance and is itself trained with VSD, allowing for the seamless integration of its components into our framework.

As shown in Fig. \ref{fig:pipeline}, the super-resolution (SR) model contains VAE encoder, SR diffusion network and VAE decoder, which are all frozen in the optimization. We denote the SR diffusion network as $\phi_{SR}$. The rendered RGB image $I$ is fed into the VAE encoder to generate latent feature $I_\mathrm{emb}$. After being added with noise, $I_\mathrm{emb}$ is denoised by the depth-to-image diffusion network $\phi_{d}$. Additionally, $I_\mathrm{emb}$ is also denoised by the SR diffusion network. The SR gradient is defined as:
\begin{equation}
\label{eq:sr}
\begin{split}
\nabla_{\theta} \mathcal{L}_{\mathrm{SR}}(\theta, d, c^{ref}) =\mathbb{E}_{t, \epsilon}[w(t)(
\epsilon_{\phi_{SR}}(\mathcal{T}(\theta); t)- \\
\epsilon_{\phi_{LoRA}}(\mathcal{T}(\theta); d, c^{ref}, t)) 
\frac{\partial \mathcal{T}(\theta)}{\partial \theta}].  
\end{split}
\end{equation}
When optimizing the parameters $\theta$ in texture $\mathcal{T}(\theta)$, the SR gradient is combined with the VSD gradient defined in Eq. \ref{eq:vsd}, and back-propagated together through the rasterizer to update $\theta$. The final gradient is defined as:
\begin{equation}
\label{eq:final}
\nabla_{\theta} \mathcal{L} = 
\nabla_{\theta} \mathcal{L}_{\mathrm{VSD}}(\theta, d, c^{ref}) + 
\lambda_{SR}\nabla_{\theta} \mathcal{L}_{\mathrm{SR}}(\theta, d, c^{ref}), 
\end{equation}
where $\lambda_{SR}$ is the weighting parameter used to balance the two gradients in optimization.
This gradient combination ensures the final texture possesses both correct large-scale structures and rich high-frequency details, yielding a sharp and visually compelling appearance.

\subsection{Texture Representation}
We use multi-resolution hash grid derived from Instant-NGP~\cite{iNGP} to represent the implicit texture $\mathcal{T}(\theta)$. In this representation, $\mathcal{T}(\theta)$ first takes UV coordinates from the rasterized renderer, quantizing these coordinates into multi-scale grid levels through a hash mapping function. The feature values from all levels are then concatenated along the feature dimension to form high-dimensional UV embeddings. The UV embeddings are then decoded into the final RGB image $I$ via a cross-attention texture decoder from \cite{SceneTex2024}. The parameters $\theta$ encompass both the parameters in texture map and the parameters in texture decoder, all of which need to be optimized.

\subsection{Implementation Details}
Our training uses 5,000 spherically distributed viewpoints sampled within the scene space and runs for 30,000 iterations. 
The learning rate is set to 0.001 in the texture field updating and 0.0001 in the LoRA module fine-tuning. We employ a time annealing strategy: uniformly sample timesteps $t\sim U(0.02, 0.98)$ for the first 5,000 iterations and then gradually shift the sampling distribution to 
$t\sim U(0.02, 0.5)$.
To ensure training stability, $\mathcal{L}_{\mathrm{SR}}$ is set to 0 for the initial 5,000 iterations and then increased to 1.2 for the remainder of the training. The whole training takes approximately 48 hours on a single NVIDIA RTX A800 GPU.
All generated textures have a resolution of $4,096 \times 4,096$. Our implementation is built upon the PyTorch, utilizing the Parameter-Efficient Fine-Tuning (PEFT) library for LoRA injection, and PyTorch3D for all differentiable rendering and texture projection.

\section{Experiment}

\subsection{Experimental Setup}

\begin{figure*}
    \centering
    \includegraphics[width=0.98\linewidth]{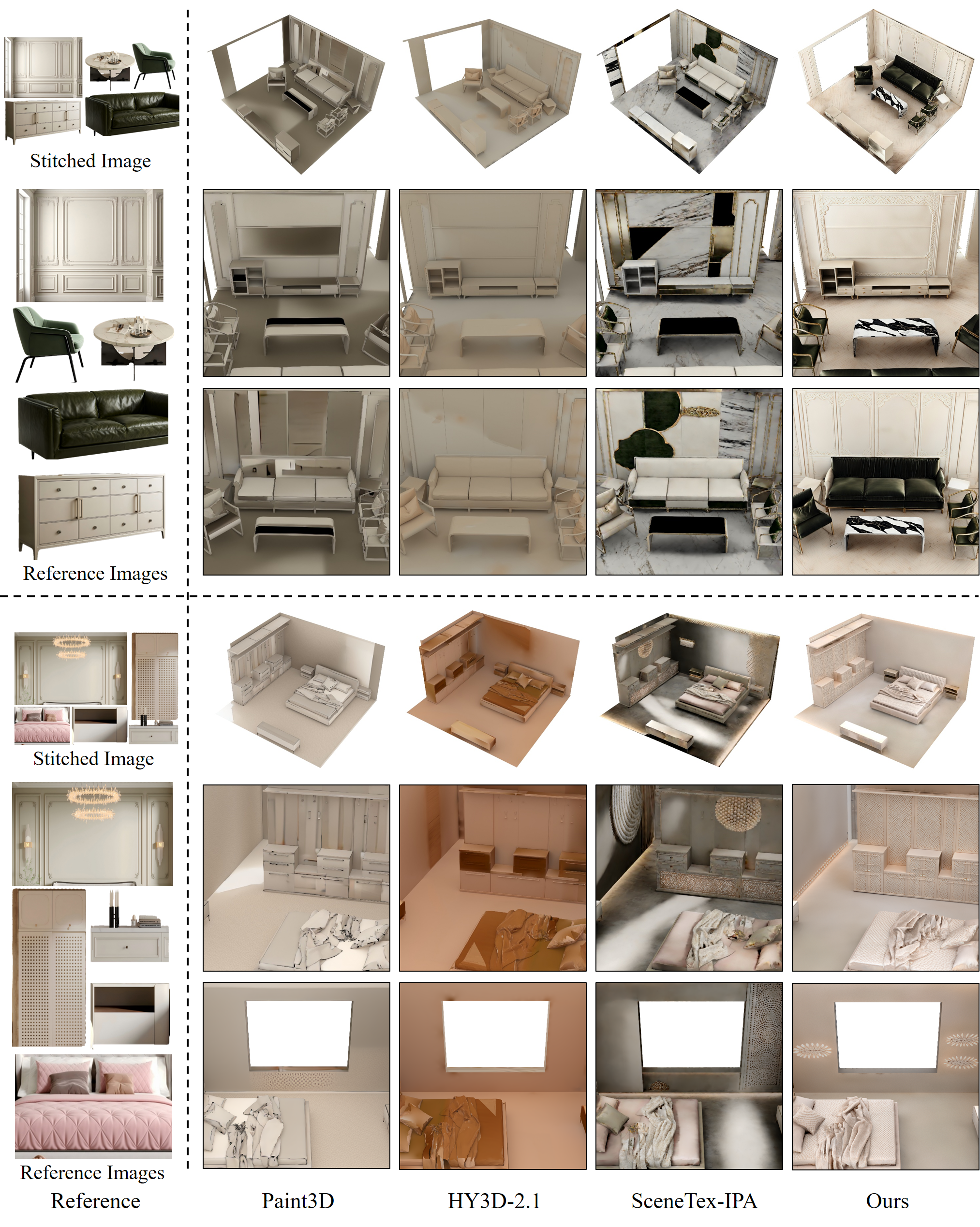}
    \caption{Qualitative comparison on image-to-texture generation. All generated textures are rendered by 3ds Max software at a resolution of $2000\times2000$ for visualization. CustomTex demonstrates instance-level consistency with the reference images, while also exhibiting greater sharpness with fewer shading effects and artifacts compared with the baselines.}
    \label{fig:img2texture}
\end{figure*}

\begin{figure}
    \centering
    \includegraphics[width=1.0\linewidth]{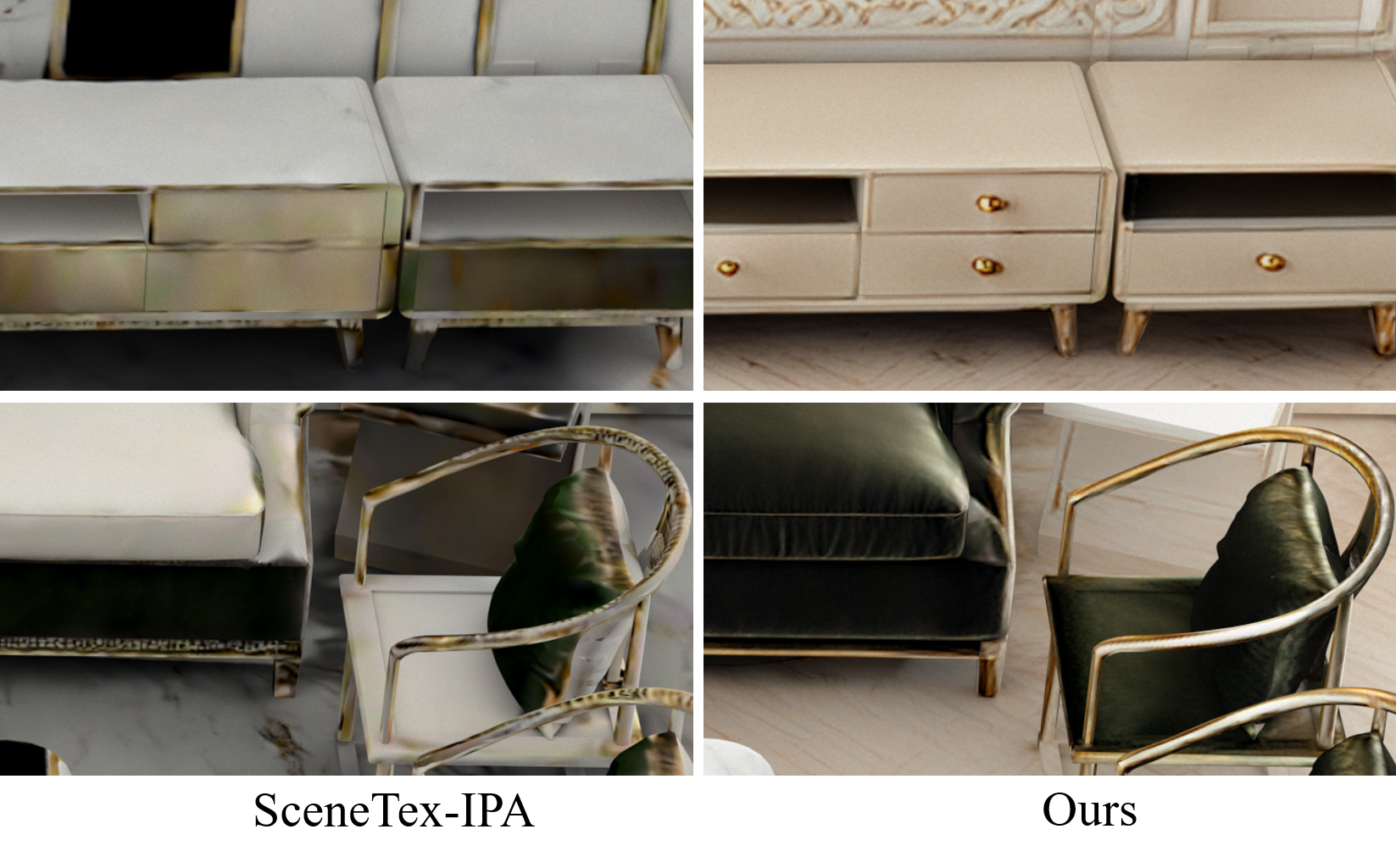}
    \caption{Qualitative comparison on close-up texture renderings.}
    \label{fig:zoomin}
\end{figure}

\begin{figure}
    \centering
    \includegraphics[width=1.0\columnwidth]{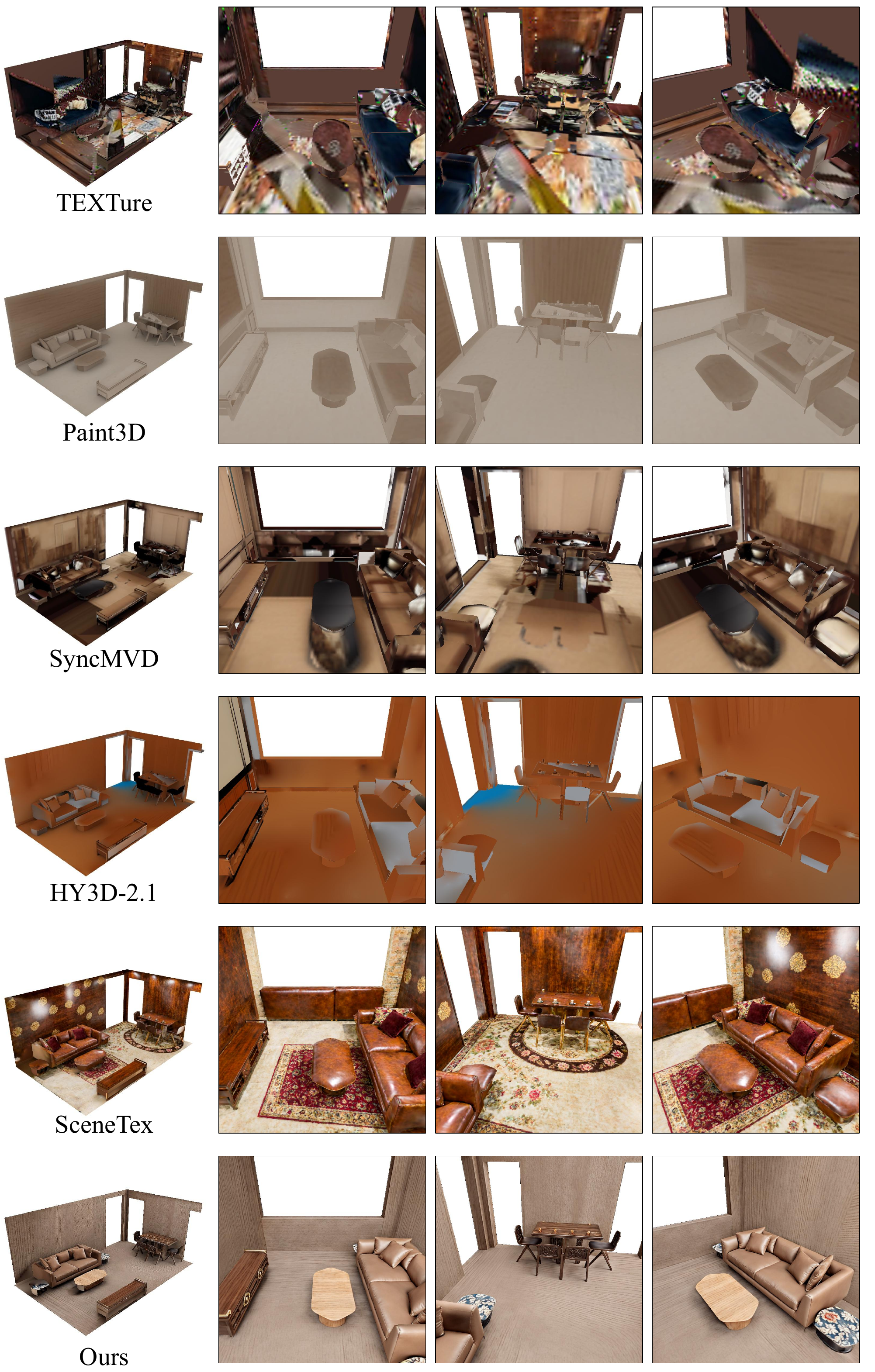}
    \caption{Qualitative comparison on text-to-texture generation. The text prompt is: ``\textit{The Nanyang vintage-style living room equipped with walls featuring dark wood panel textures, a brown leather sofa, a round fabric stool with floral patterns, a TV stand made of dark wood with golden handles, dark brown wooden chairs and a light-color wood coffee table.}" GPT-4v is used to convert this text prompt into reference image prompts for our CustomTex. All generated textures are rendered to $768\times768$ resolution images for visualization. Only CustomTex demonstrates instance-level consistency with the text prompt.}
    \label{fig:tex2texture}
    \vspace{-10pt}
\end{figure}

\textbf{Baselines.} 
We compare CustomTex against a comprehensive set of texture synthesis baselines to demonstrate its effectiveness. These baselines include image-to-texture methods like Paint3D~\cite{Paint3D}, HY3D-2.1~\cite{HY3D-2.1} and SceneTex-IPA~\cite{SceneTex2024}, and text-to-texture methods like TEXture~\cite{TEXTure}, Paint3D~\cite{Paint3D}, SyncMVD~\cite{SyncMVD}, HY3D-2.1~\cite{HY3D-2.1} and SceneTex~\cite{SceneTex2024}. Paint3D and HY3D-2.1 support both image and text prompts. As the original SceneTex is limited to text prompt, we adapt it by integrating an IP-Adapter~\cite{IPAdapter} to accommodate image prompt and name this variant as SceneTex-IPA. Since all image-to-texture baselines don't support multiple image prompts, we stitch the reference images into a single large image to serve as the image prompt for these baselines. 
As for the text-to-texture comparison, we utilize GPT-4v to generate the set of reference images for our CustomTex from a fine-grained textual prompt.

\textbf{Evaluation Metrics.} We employ a multi-faceted evaluation protocol tailored to the two comparison scenarios. For comparison with the image-to-texture baselines, we adopt CLIP-Score (CLIP-I)~\cite{CLIP} and CLIP-FID (a CLIP version of FID~\cite{FID}) to quantify the distributional similarity between renderings of the textured meshes and the reference images. Furthermore, to incorporate a human-centric perspective, we leverage the latest LMM-based model Q-Align~\cite{Qalign} for comprehensive Image Quality Assessment (Q-Align IQA) and Image Aesthetic Assessment (Q-Align IAA) of the final textured mesh renderings. For comparison with text-to-texture baselines, we follow SceneTex~\cite{SceneTex2024} that utilizes CLIP-Score (CLIP-T) and Inception Score (IS)~\cite{voxel1} to measure semantic fidelity to the input prompts and generated texture quality respectively. We conduct quantitative experiments on the same evaluation dataset as SceneTex~\cite{SceneTex2024}, which comprises 10 scenes sampled from 3D-FRONT~\cite{3D-FRONT2021}.


\subsection{Method Comparison}
\label{sec:method_comparison}


\textbf{Quantitative Results.} 
Our CustomTex demonstrates quantitative superiority in both image-to-texture and text-to-texture generation. As shown in Tab.~\ref{tab:img2texture}, CustomTex achieves top performance across all four metrics in the image-to-texture task. Its leading CLIP-I and CLIP-FID scores indicate that the textures produced by CustomTex exhibit greater distributional similarity to the reference images, and its leading Q-Align IQA and IAA scores  reflect its higher-quality texture generation.
This superiority extends to the text-to-texture task. As evidenced by Tab. \ref{tab:tex2texture}, CustomTex achieves the highest scores across all metrics, demonstrating strong semantic fidelity to text prompts (CLIP-T) while also attaining superior image quality (IS, Q-Align IQA and Q-Align IAA).

\begin{table}
\centering
\footnotesize

\setcellgapes{2pt} 
\makegapedcells
\begin{tabular}{l|cccc}
\hline
\multirow{2}{*}{\raggedright Method}  & \multirow{2}{*}{CLIP-I$\uparrow$} & \multirow{2}{*}{CLIP-FID$\downarrow$}  & {Q-Align} & {Q-Align}
\\ 
{}  & {} & {}  & {IQA$\uparrow$} & {IAA$\uparrow$}
\\
\hline
{Paint3D~\cite{Paint3D}}   & {0.694}  & {130.138} & {2.896}  & {2.401} \\
{HY3D-2.1~\cite{HY3D-2.1}}   & {0.682}  & {134.680} & {2.187}  & {1.838} \\
{SceneTex-IPA~\cite{SceneTex2024}}   & {0.741}  & {121.118} & {4.009}  & {3.594} \\
\hline
{CustomTex (Ours)}  & \textbf{0.797}  & \textbf{106.229} & \textbf{4.469} & \textbf{3.629} \\
\hline
\end{tabular}

\caption{Quantitative comparison on image-to-texture generation.} 
\label{tab:img2texture}
\vspace{-10pt}
\end{table}

\begin{table}
\centering
\footnotesize
\setcellgapes{2pt} 
\makegapedcells
\begin{tabular}{l|cccc}
\hline
\multirow{2}{*}{\makecell[l]{Method}} & \multirow{2}{*}{CLIP-T$\uparrow$} & \multirow{2}{*}{IS$\uparrow$}  & {Q-Align} & {Q-Align}
\\ 
{}  & {} & {}  & {IQA$\uparrow$} & {IAA$\uparrow$}
\\
\hline
{TEXture~\cite{TEXTure}} & {0.557} & {1.372}  & {1.645} & {1.574}\\
{Paint3D~\cite{Paint3D}}   & {0.734}  & {2.330} & {2.442}  & {1.792} \\
{SyncMVD~\cite{SyncMVD}}   & {0.712}  & {2.467} & {2.409}  & {2.328} \\
{HY3D-2.1~\cite{HY3D-2.1}}   & {0.734}  & {2.381} & {2.774}  & {2.033} \\
{SceneTex~\cite{SceneTex2024}}   & {0.639}  & {3.009} & {3.824}  & {2.681} \\
\hline
{CustomTex (Ours)}  & \textbf{0.766}  & \textbf{3.311} & \textbf{4.252} & \textbf{3.343}\\
\hline
\end{tabular}
\caption{Quantitative comparison on text-to-texture generation.} 
\label{tab:tex2texture}
\vspace{-10pt}
\end{table}

\textbf{Qualitative Results.} 
Fig. \ref{fig:img2texture} presents the qualitative comparison on image-to-texture generation.
Paint3D~\cite{Paint3D} and HY3D-2.1~\cite{HY3D-2.1} 
fail to correctly interpret stitched reference image prompts, resulting in repetitive patterns and an unnaturally uniform style. SceneTex-IPA~\cite{SceneTex2024} achieves greater similarity to the reference yet still deviates from it. In contrast, our CustomTex demonstrates precise instance-level consistency with the reference across objects like sofa, chair, cabinet, tea table and walls. Additionally, CustomTex produces textures with enhanced sharpness and a notable reduction in shading effects and artifacts compared to SceneTex-IPA~\cite{SceneTex2024}. Fig. \ref{fig:zoomin} presents close-up visual comparisons of the rendered textures, highlighting differences in quality, sharpness and detail.

Fig. \ref{fig:tex2texture} presents the qualitative comparison on text-to-texture generation. The text prompt specifies fine-grained attributes for each object in the scene. However, TEXture~\cite{TEXTure}, Paint3D~\cite{Paint3D}, SyncMVD~\cite{SyncMVD} and HY3D-2.1~\cite{HY3D-2.1} fail to correctly interpret this complex text prompt, resulting in visually irrational and low-quality textures.
SceneTex~\cite{SceneTex2024} produces more globally reasonable and natural textures, but the textures of the walls and coffee table are not consistent with the text prompt, which specifies ``walls featuring dark wood panel textures" and ``light-color wood coffee table." CustomTex uses image prompt to produce textures with precise instance-level consistency to the complex text prompt. This comparison indicates that text prompt struggles to convey precise visual characteristics compared with image prompt.

\begin{table}
\centering
\footnotesize

\setcellgapes{2pt} 
\makegapedcells
\begin{tabular}{l|cc}
\hline
{Method} & {Visual Quality$\uparrow$} & {Prompts Consistency$\uparrow$}
\\
\hline

{Paint3D~\cite{Paint3D}}   & {2.792}  & {2.817} \\

{HY3D-2.1~\cite{HY3D-2.1}}   & {2.525}  & {2.619} \\
{SceneTex-IPA~\cite{SceneTex2024}}   & {3.842}  & {3.617}\\
\hline
{CustomTex (Ours)}  & \textbf{4.008}  & \textbf{4.125} \\
\hline
\end{tabular}

\caption{User study results on visual quality and prompt consistency, averaged over 60 participants on a 1–5 scale. Our method achieves the highest scores across both criteria.} 
\label{tab:user_study}
\vspace{-5pt}
\end{table}

\textbf{User Study.} 
We conducted a user study to further evaluate CustomTex in terms of perceived visual quality and consistency. A total of 60 participants rated each generated texture based on visual quality, including clarity, color and composition, as well as the consistency between the texture content and the image prompt. Ratings were provided on a 1-to-5 scale, where 1 indicates ``very poor" or ``completely inconsistent", and 5 indicates ``very good" or ``highly consistent".
The results reported in Tab. \ref{tab:user_study} indicate that our method is consistently preferred.  



\begin{figure}
    \centering
    \includegraphics[width=0.98\linewidth]{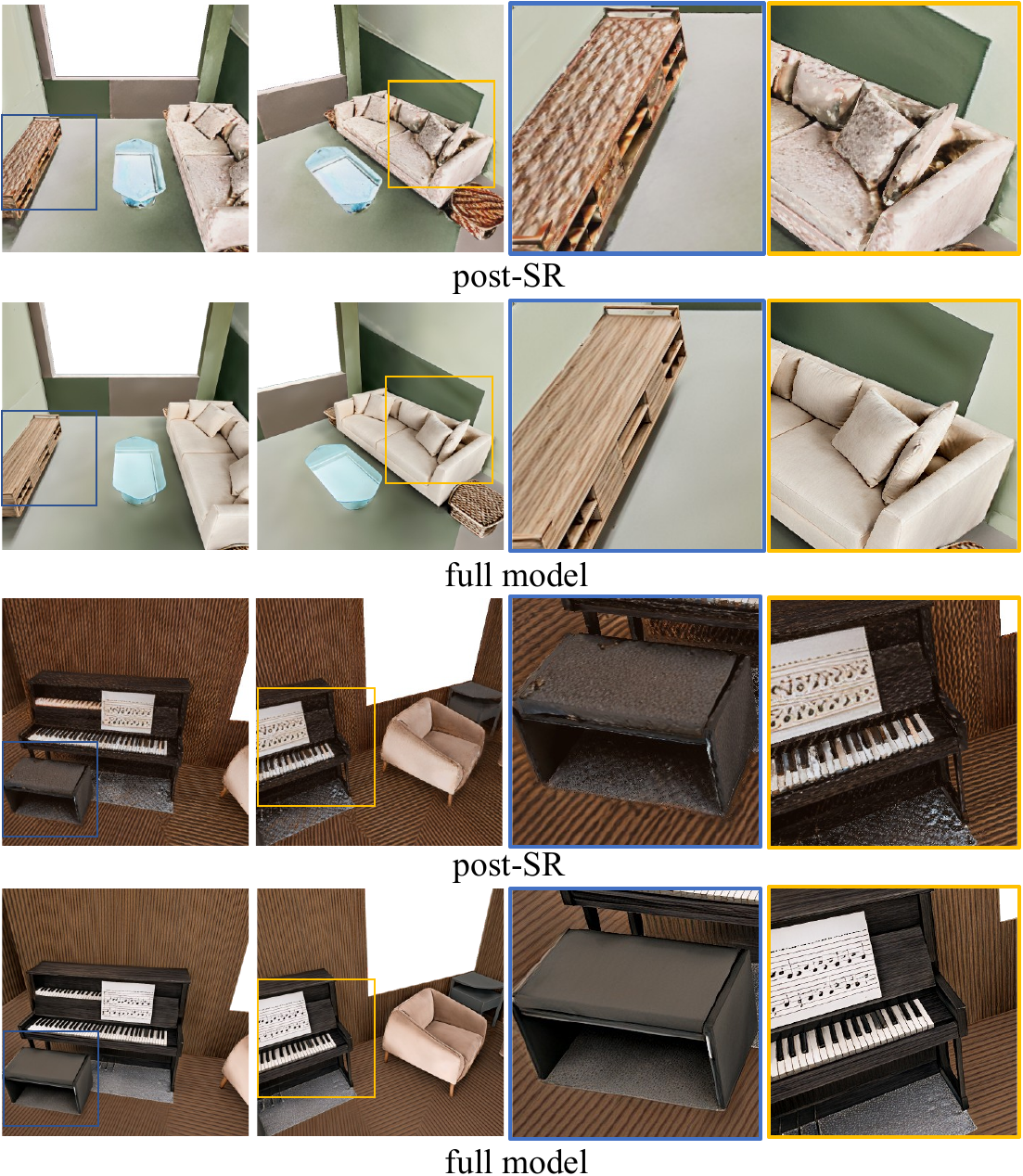}
    \caption{Qualitative ablation study results: \textit{post-SR} vs. \textit{full model}.
    }
    \label{fig:postsr}
\end{figure}

\begin{figure*}
    \centering
    \includegraphics[width=1\linewidth]{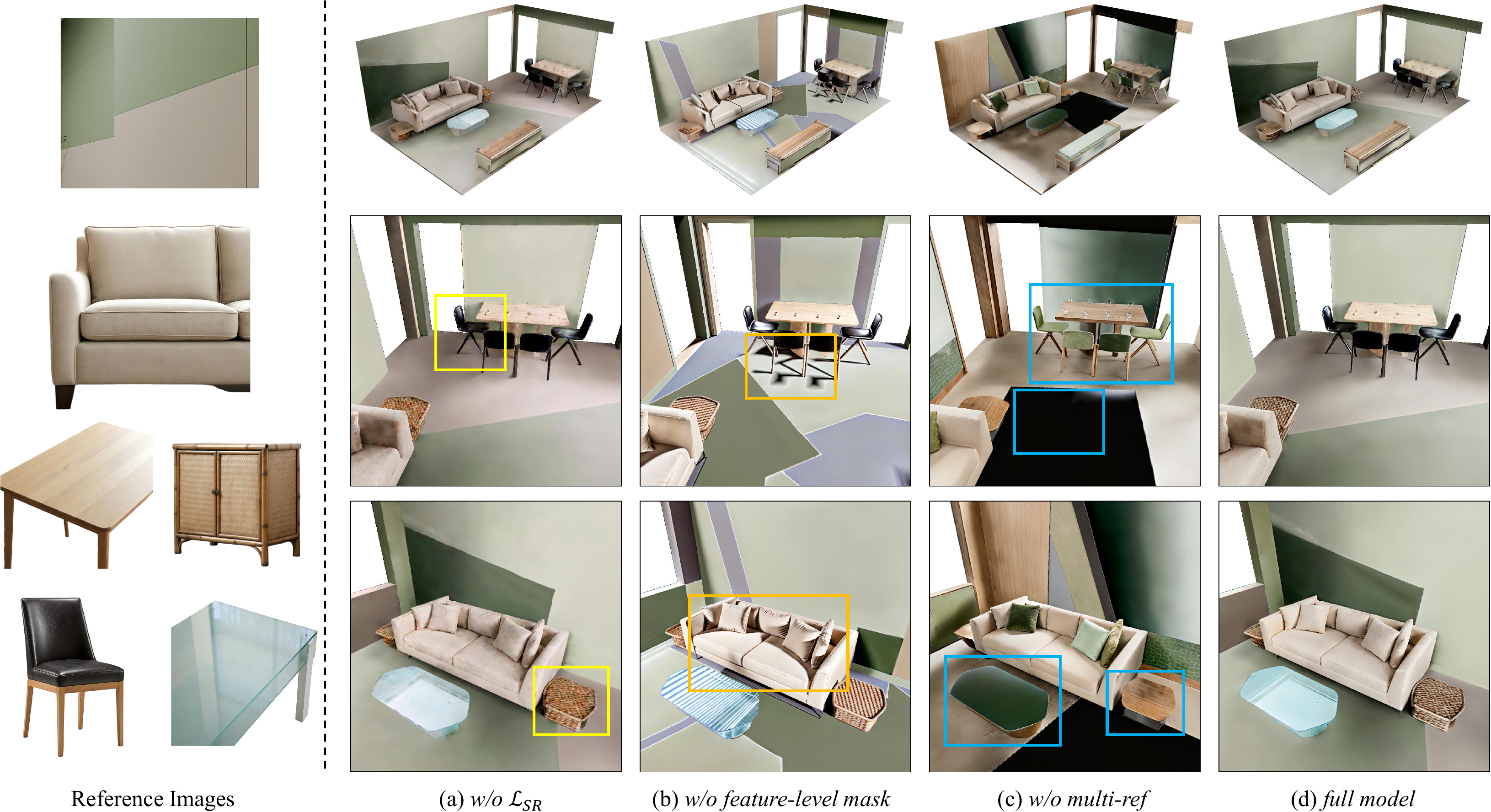}
    \caption{Qualitative ablation study results.}
    \label{fig:abaltion_studies}
    \vspace{-10pt}
\end{figure*}

\subsection{Ablation Study}

\begin{table}
\centering
\footnotesize

\setcellgapes{2pt} 
\makegapedcells
\begin{tabular}{l|cccc}
\hline
\multirow{2}{*}{\raggedright Method}  & \multirow{2}{*}{CLIP-I$\uparrow$} & \multirow{2}{*}{CLIP-FID$\downarrow$}  & {Q-Align} & {Q-Align}
\\ 
{}  & {} & {}  & {IQA$\uparrow$} & {IAA$\uparrow$}
\\
\hline
post-SR & {0.746}  & {114.612} & {2.959}  & {2.190} \\
w/o $\mathcal{L}_{\mathrm{SR}}$   & {0.736}  & {118.247} & {3.330}  & {2.664} \\
w/o multi-ref & {0.757}  & {109.243} & {4.053}  & {3.519} \\
w/o f-mask  & {0.743}  & {111.205} & {3.689}  & {3.231} \\
\hline
full model  & \textbf{0.797}  & \textbf{106.229} & \textbf{4.469} & \textbf{3.629} \\
\hline
\end{tabular}
\caption{Quantitative ablation study results. We validate the effectiveness of our method by testing four model variants: \textit{post-SR}, w/o \textit{$\mathcal{L}_{\mathrm{SR}}$}, w/o \textit{feature-level mask} and w/o\textit{ multi-ref}, showcasing their distinct contributions.} 
\label{tab:ablation_studies}
\vspace{-5pt}
\end{table}

To comprehensively analyze the contribution of each module, we evaluate four variants of CustomTex: \textit{post-SR}, w/o \textit{$\mathcal{L}_{\mathrm{SR}}$}, w/o \textit{feature-level mask} and w/o\textit{ multi-ref}. The qualitative ablation results are shown in Fig.~\ref{fig:postsr} and Fig.~\ref{fig:abaltion_studies}. The quantitative ablation results are reported in Tab.~\ref{tab:ablation_studies}.

\textbf{Can we directly conduct image super-resolution to the generated textures as a post-processing step?}
To investigate this, we evaluate an important variant of our proposed CustomTex, which does not incorporate pixel-level distillation during optimization, but performs image super-resolution \cite{PiSA-SR2025} on the generated textures as a post-processing step. We denote this variant as \textit{post-SR}. The quantitative evaluation results presented in Tab. \ref{tab:ablation_studies} indicate that, this variant leads to a significant drop in two image quality assessment metrics (Q-Align IQA and IAA). This finding is further corroborated by the qualitative comparison in Fig. \ref{fig:postsr},
which shows that the textures produced by \textit{post-SR} contain obvious blurriness, noise and artifacts, whereas our method yields better sharpness and clarity. Hence, we conclude that incorporating the image super-resolution model into the distillation process is more effective for improving texture quality than using it as a post-processing step.

\begin{figure}
    \centering
    \includegraphics[width=0.95\linewidth]{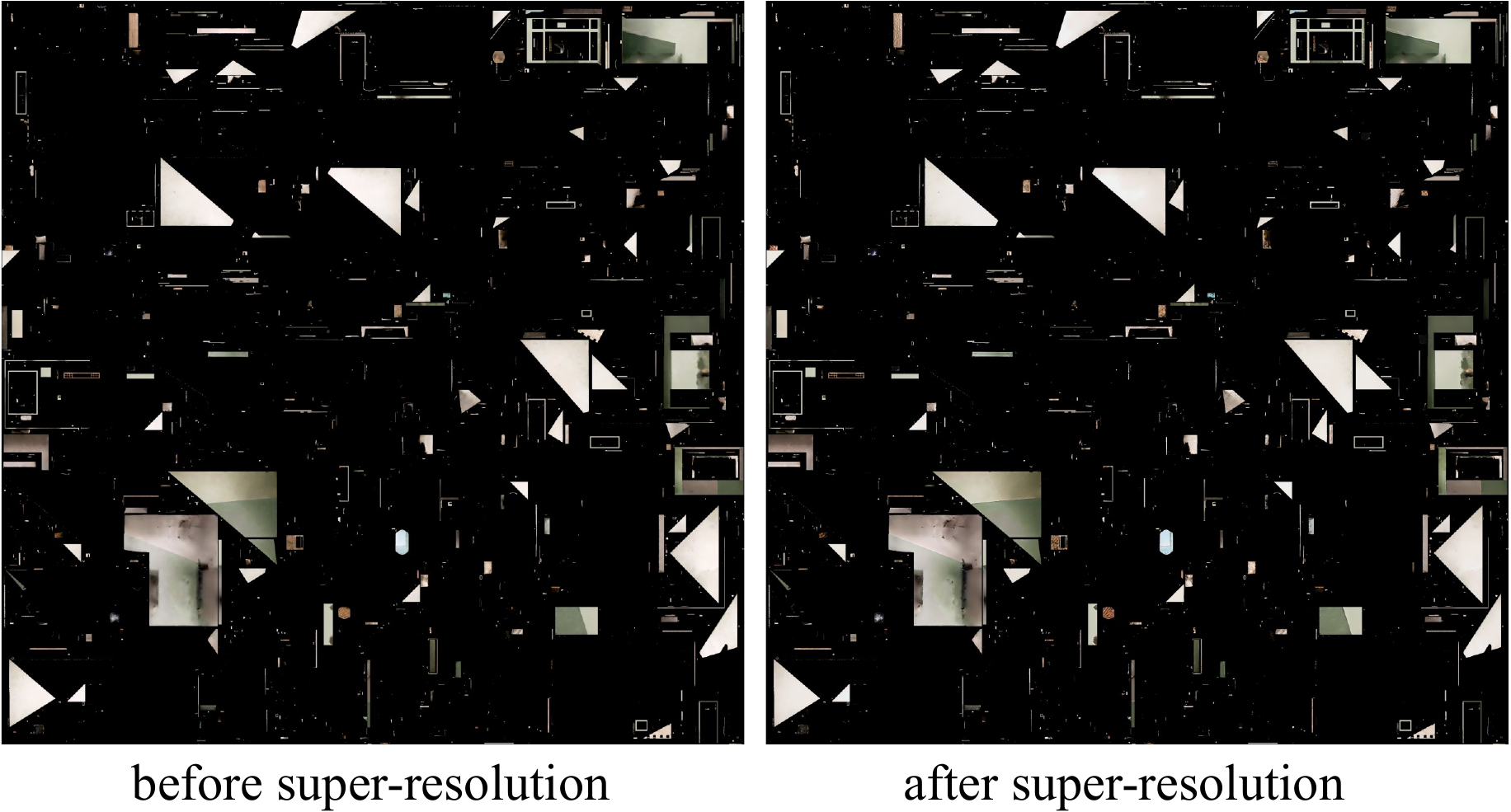}
    \caption{Textures before and after image super-resolution operation. These two textures have a resolution of $4,096\times4,096$, and correspond to the scene shown in the first row of Fig. \ref{fig:postsr}.
    Black pixels in textures denote invalid regions where UV coordinates do not map to any vertex in the mesh.
    }
    \label{fig:beforesr}
\end{figure}

Fig. \ref{fig:beforesr} compares the UV textures before and after the super-resolution operation, demonstrating that the process does not improve the texture quality. This is because UV textures, unlike the rendered images in Fig. \ref{fig:postsr}, typically lack the high-level and middle-level semantic structures (e.g., distinct objects, object parts, or surface textures) found in natural images. Since most super-resolution models were trained on natural images, they are ill-suited for UV textures and fail to produce satisfactory results when applied directly.

Another potential super-resolution approach is to apply the process to the rendered images rather than the textures themselves. While this can enhance the immediate visual output, it does not improve the intrinsic quality of the textures. Acquiring high-quality textures is essential, as many downstream applications rely on them as fundamental 3D assets alongside the mesh.

\textbf{Does pixel-level distillation truly improve texture quality?} To answer this, we let $\lambda_{SR}$ equal 0 in the Eq. \ref{eq:final} and conduct only semantic-level distillation, under the \textit{w/o $\mathcal{L}_{\mathrm{SR}}$} setting.
As shown in Tab.~\ref{tab:ablation_studies}, 
this leads to a significant drop in two image quality assessment metrics (Q-Align IQA and IAA).
Fig.~\ref{fig:abaltion_studies} (a) further reveals that the results exhibit increased blurriness and artifacts. This confirms that pixel-level distillation provides crucial high-fidelity supervision, which helps preserve finer textural details and sharpness.

\textbf{Is feature-level masking an effective way to fuse reference images and instance masks?} In the \textit{w/o feature-level mask} (f-mask) configuration, we apply instance masks at the noise-level instead of the feature-level in cross-attention layers, which can be represented by the formula:
$\epsilon_{\phi_d} = \frac{1}{N} \sum_{i=1}^N {m_i} \epsilon_{\phi_d}(\mathcal{T}(\theta); d, c^{ref}_{i}, t)$, and the same operation is also applied to $\epsilon_\phi$. 
As shown in Fig.~\ref{fig:abaltion_studies} (b), this causes unstable lighting around objects, indicating that feature-level masking better aligns the instance cues with the generated features while more effectively preserving lighting stability.

\textbf{What are the effects of merging all reference images into a single large image instead of using multi-reference inputs?}
In the \textit{w/o multi-ref} setting, we concatenate all reference images (e.g., sofa, bed, table, etc.) into one large composite input. As shown in Fig.~\ref{fig:abaltion_studies} (c), this ablated setting makes it difficult for the model to distinguish between object instances, resulting in notable inconsistency between the reference images and the generated targets. These results highlight the necessity of maintaining multi-reference input for clear instance separation and fine-grained controllability.


\textbf{Why can our method produce textures with less ``baked-in” shading?}  This is primarily attributed to the multi-reference conditioning mechanism. As shown in Fig. \ref{fig:abaltion_studies} (c), \textit{w/o multi-ref} introduces obvious shading on the generated textures, such as walls and floors. The Stable Diffusion model often introduces strong, global shading to enhance perceived realism. By using instance masks, our method decomposes the global image generation process into the generation of local object appearances, thereby preventing the formation of overarching shading across the entire image.

\begin{figure}
    \centering
    \includegraphics[width=0.9\linewidth]{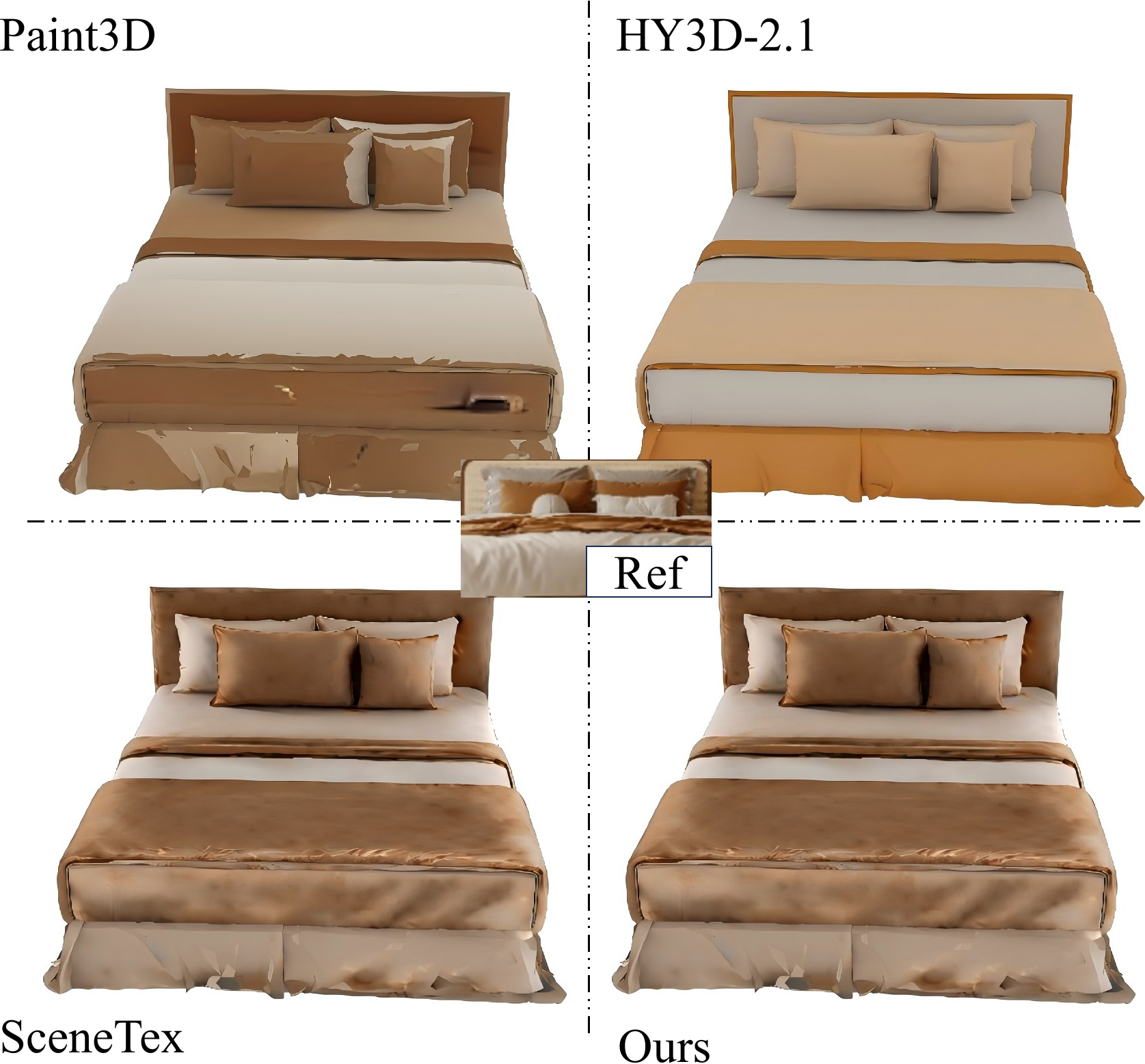}
    \caption{Comparison on single-object texturing.}
    \label{fig:single_objs}
\end{figure}


\subsection{Additional Comparison}

\textbf{Comparison on single-object texturing}. In Fig. \ref{fig:single_objs}, we show the comparison on single-object texture generation (without image stitching). The style of the texture generated by our method is still more consistent with the reference. Paint3D~\cite{Paint3D} and HY3D-2.1~\cite{HY3D-2.1} perform well in single-object texturing but struggles with scene-level texturing.

\begin{figure*}
    \centering
    \includegraphics[width=0.99\linewidth]{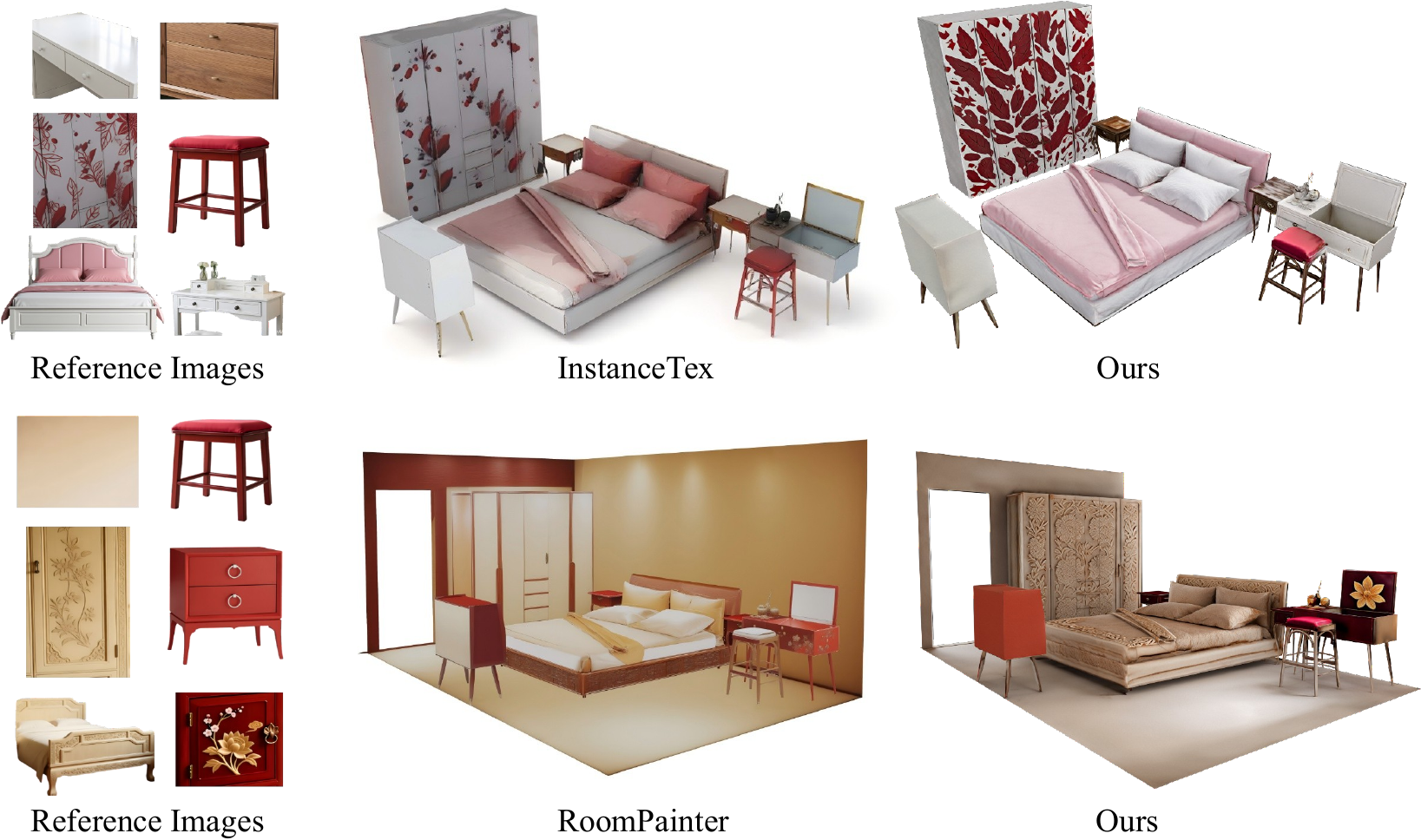}
    \caption{Visual comparison with InstanceTex~\cite{InstanceTex2024} and RoomPainter~\cite{scene2}.}
    \label{fig:more_compare}
\end{figure*}

\textbf{Comparison with closed-source methods}. In Sect.~\ref{sec:method_comparison}, we have compared our method against existing texture synthesis methods for which code or pre-trained models are publicly available. We don't include two most recent works, InstanceTex~\cite{InstanceTex2024} and RoomPainter~\cite{scene2}, as their implementations have not been released. To enable visual comparison, we reproduce their results using the images presented in their original papers and retrieve the most visually similar object images from online repositories to as reference images for our method. As shown in Fig. \ref{fig:more_compare}, the textures generated by our method exhibit greater visual richness and realism than those produced by the two competitors.


\subsection{Inference Time}

By leveraging a multi-resolution hash grid for implicit texture representation, our method generates texture maps at arbitrary resolutions. We further enhance inference efficiency through a patch-based texture generation strategy. As a result, our method produces textures up to 4K within seconds and maintains high efficiency even at 12K resolution, requiring only about 20 seconds. Complete inference speeds across various resolutions are detailed in Tab.~\ref{tab:inference_time}.

\begin{table}
    \centering
    \footnotesize
    
\setcellgapes{3pt} 
\makegapedcells
\begin{tabular}{l|c}
\hline
{Texture resolution} & {Inference time (s)}
\\
\hline
{$1,024\times1,024$} & {0.15} \\
\hline
{$2,048\times2,048$} & {0.54} \\
\hline
{$3,072\times3,072$} & {1.19} \\
\hline
{$4,096\times4,096$} & {2.41} \\
\hline
{$8,192\times8,192$} & {9.69} \\
\hline
{$12,288\times12,288$} & {21.69} \\
\hline
\end{tabular}
    \caption{Inference time across different texture resolutions} 
    \label{tab:inference_time}
\end{table}

\begin{figure*}
    \centering
    \includegraphics[width=1\linewidth]{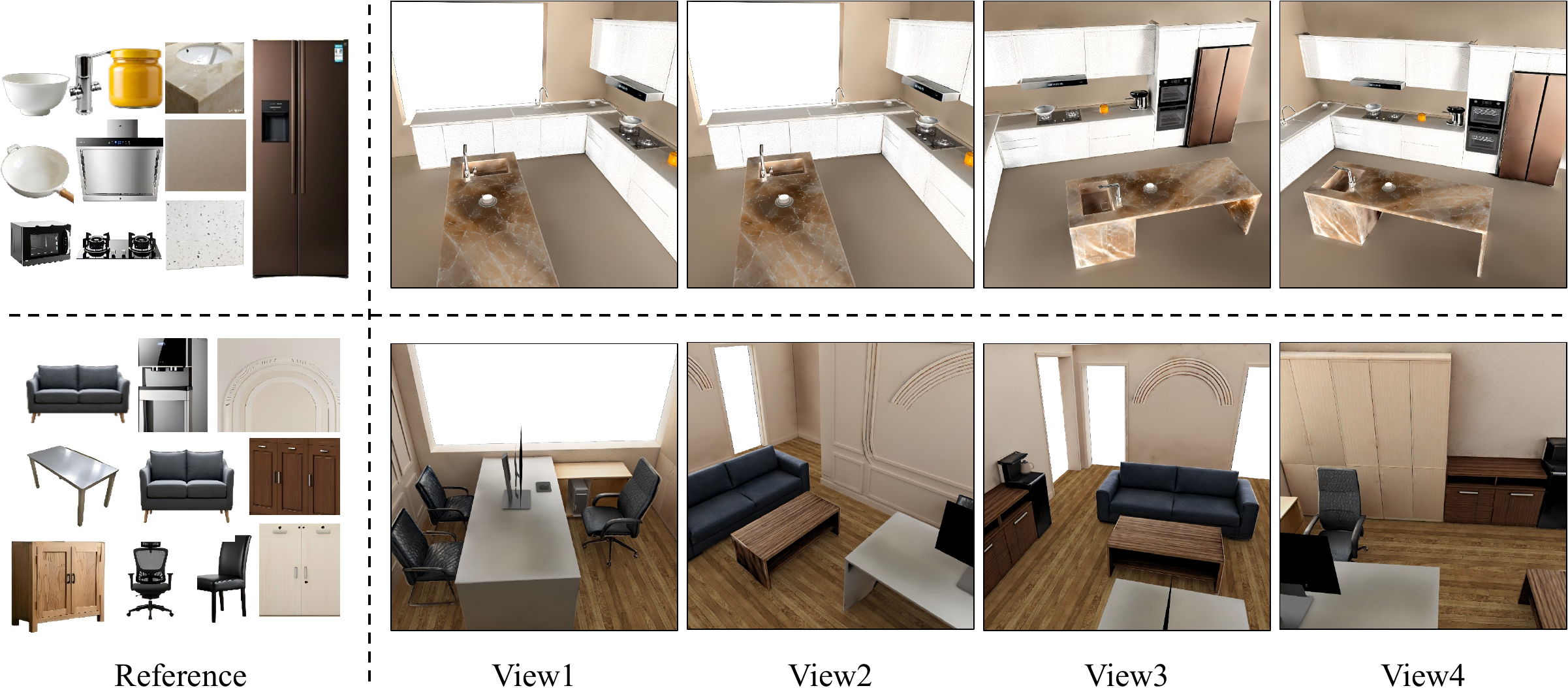}
    \caption{Qualitative results on challenging scenes with diverse room types and complex layouts.}
    \label{fig:challenge_case}
    \vspace{-10pt}
\end{figure*}

\section{More Visual Results}
\begin{figure*}
    \centering
    \includegraphics[width=0.99\linewidth]{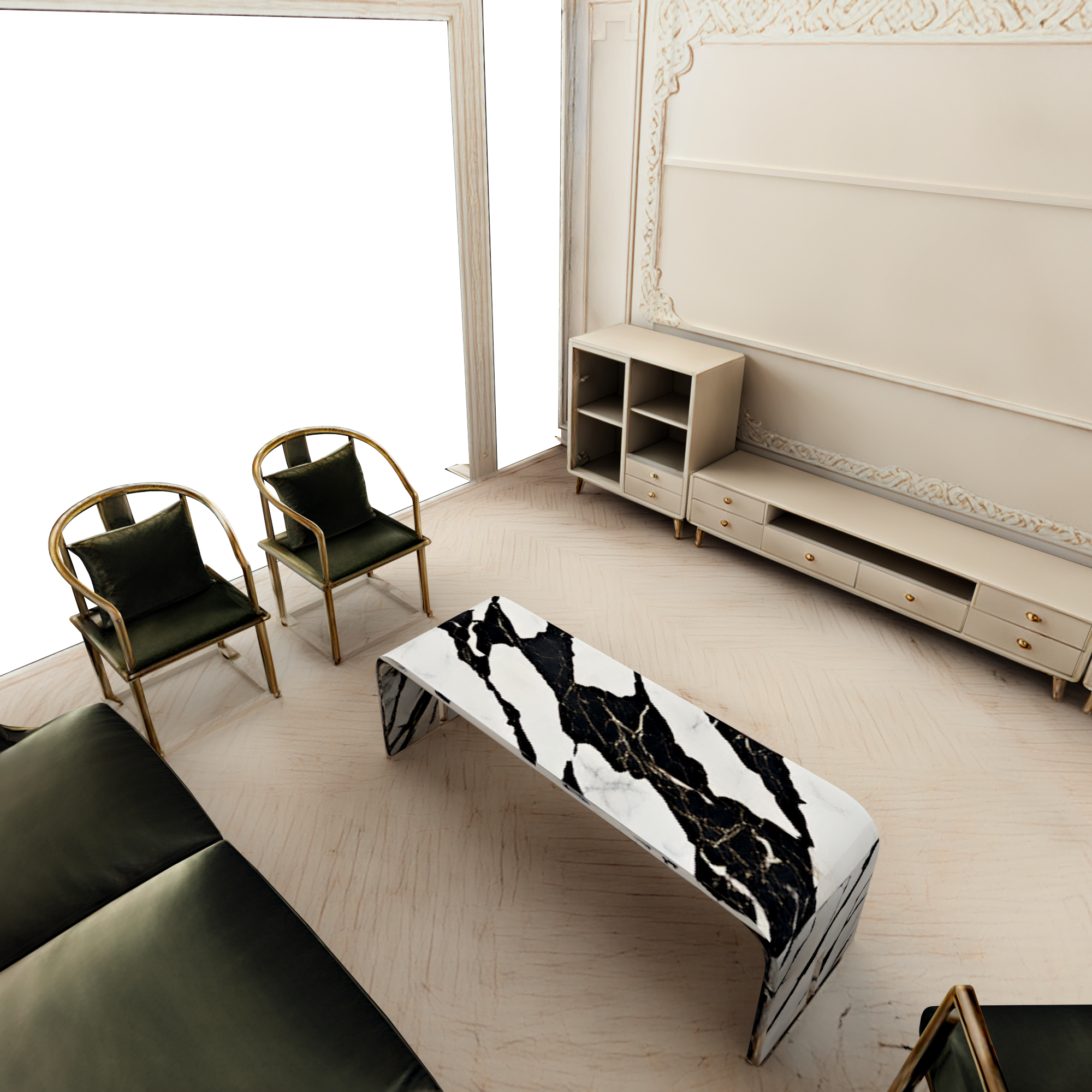}
    \caption{The ``living room" texture generated by our method is rendered into $2,000\times2,000$ resolution image.
    }
    \label{fig:bigimage1}
\end{figure*}

\begin{figure*}
    \centering
    \includegraphics[width=0.99\linewidth]{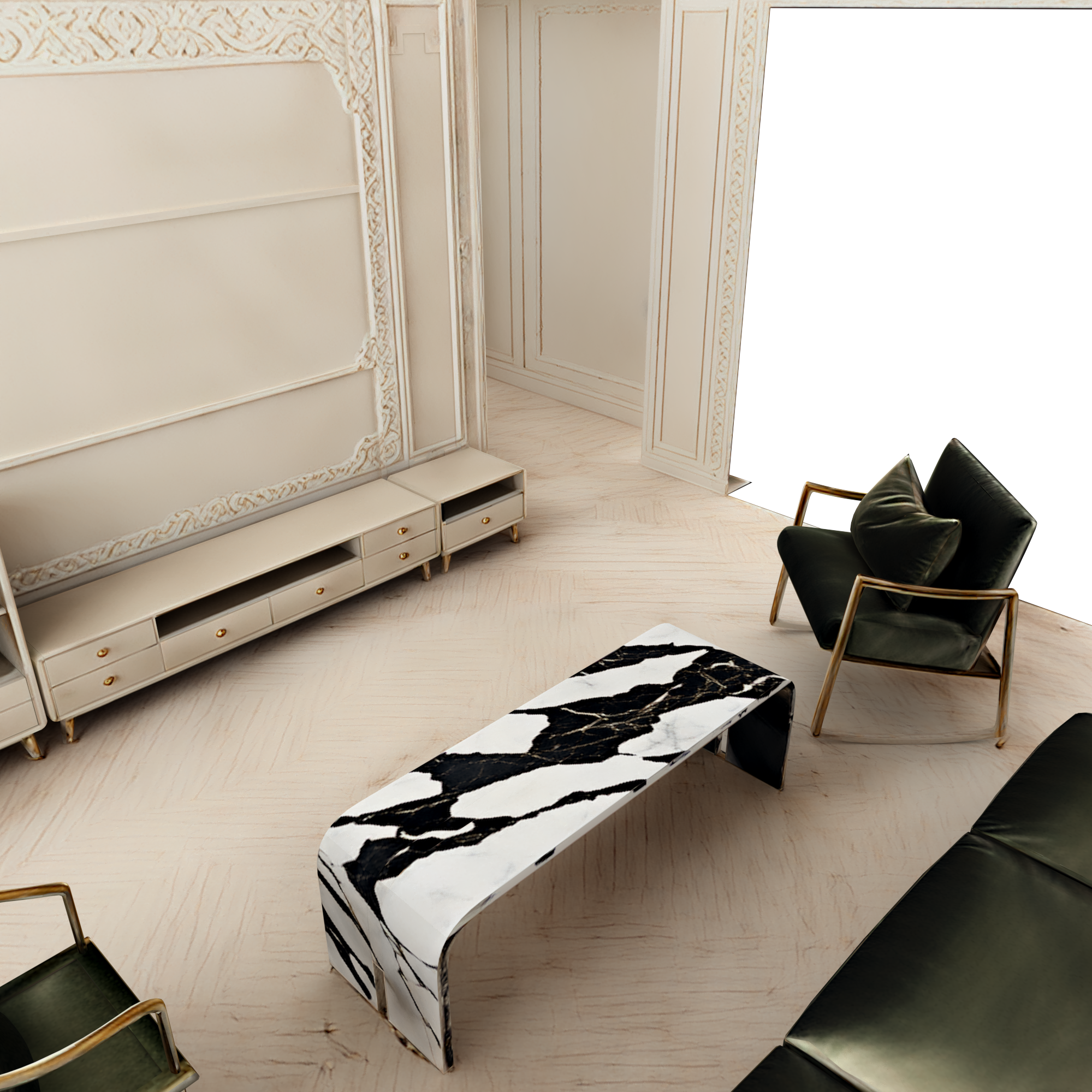}
    \caption{The ``living room" texture generated by our method is rendered into $2,000\times2,000$ resolution image.
    }
    \label{fig:bigimage2}
\end{figure*}

\begin{figure*}
    \centering
    \includegraphics[width=0.99\linewidth]{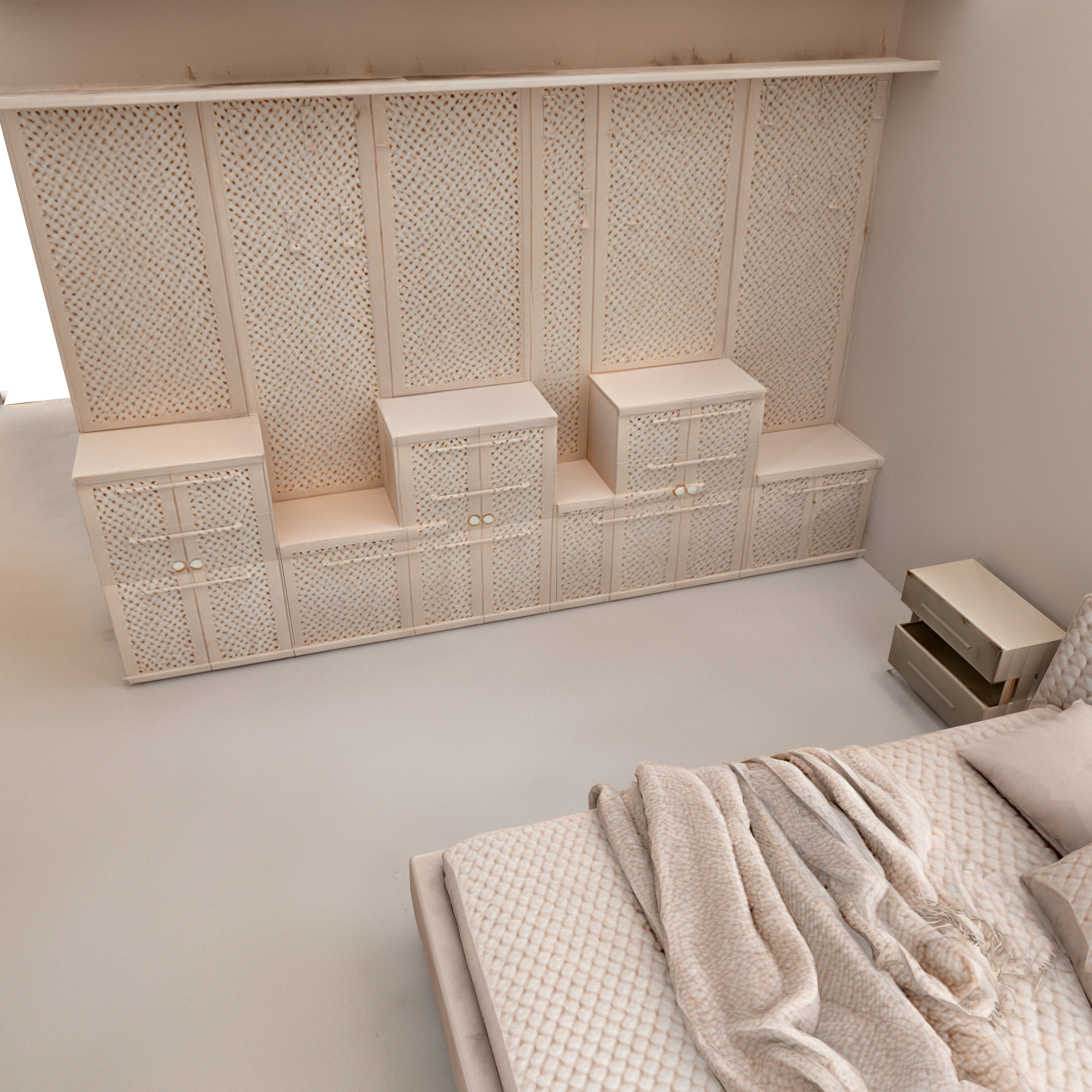}
    \caption{The ``bedroom" texture generated by our method is rendered into $2,000\times2,000$ resolution image.
    }
    \label{fig:bigimage3}
\end{figure*}

\begin{figure*}
    \centering
    \includegraphics[width=0.99\linewidth]{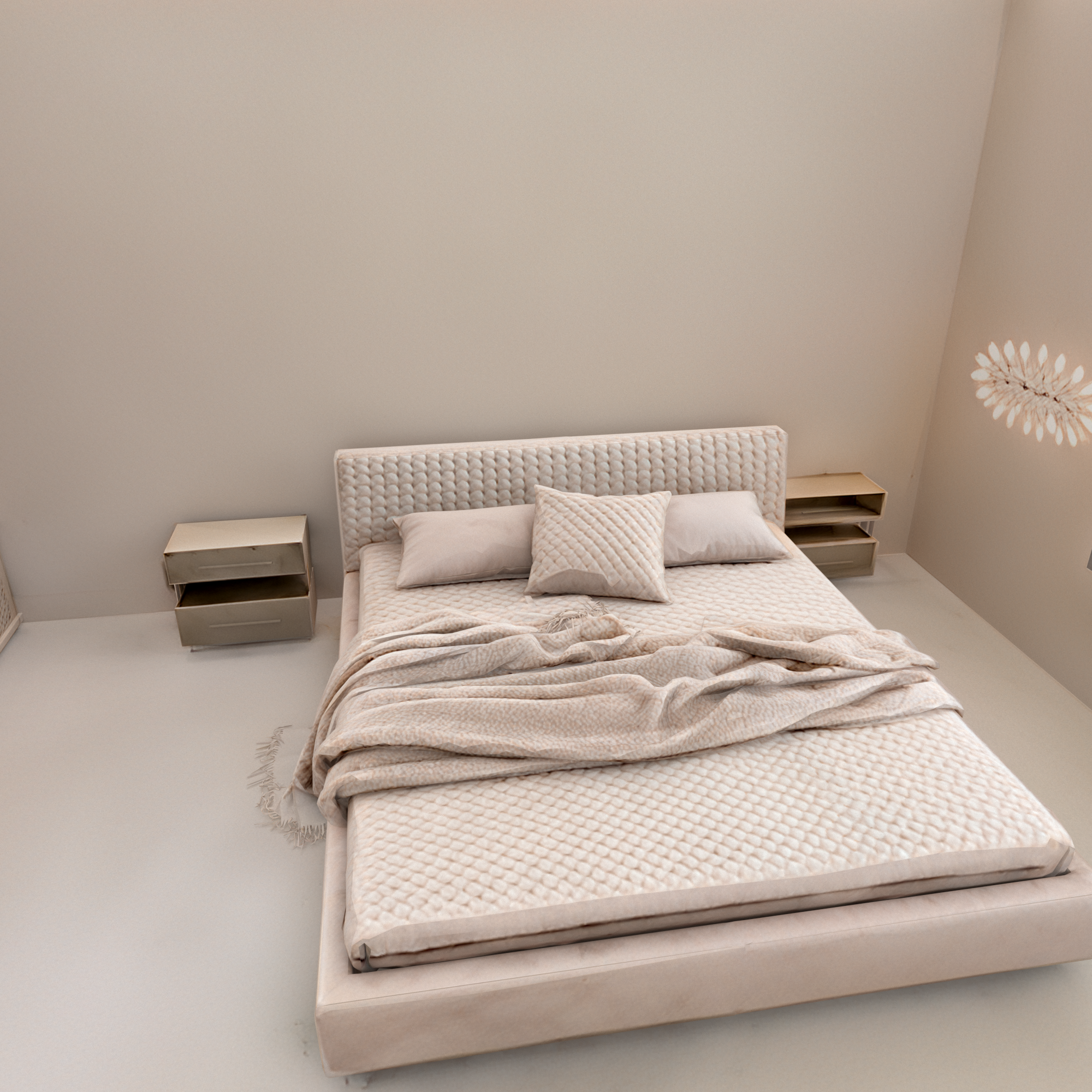}
    \caption{The ``bedroom" texture generated by our method is rendered into $2,000\times2,000$ resolution image.
    }
    \label{fig:bigimage4}
\end{figure*}

\begin{figure*}
    \centering
    \includegraphics[width=0.99\linewidth]{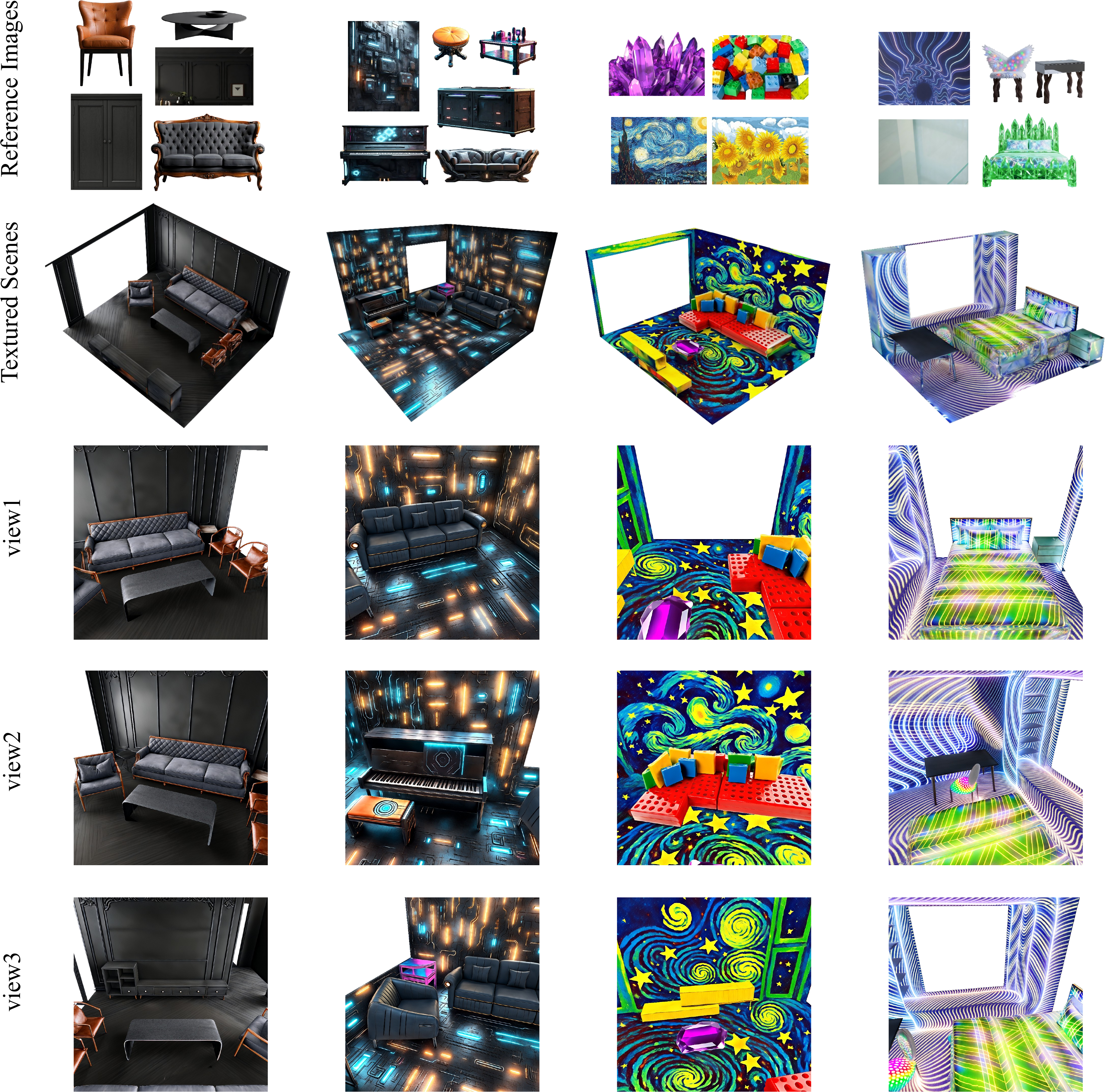}
    \caption{Using images with more diverse styles as the reference images.
    }
    \label{fig:styleimage}
\end{figure*}

Fig.~\ref{fig:challenge_case} presents visual results produced by our method on two challenging scenes, a kitchen and an office, which feature diverse layouts and materials. These results indicate that our method performs well in complex scene texturing.
Fig. \ref{fig:bigimage1}-\ref{fig:bigimage4} show the $2,000\times2,000$ resolution images rendered from our generated textures in two scenes ``living room'' and ``bedroom''. Our method produces visually compelling textures with significantly reduced blurriness, artifacts and ``baked-in'' shading. Fig. \ref{fig:styleimage} shows that our method successfully generalizes to various exaggerated styles beyond furniture. The results, generated from reference images including ``dark'', ``Van Gogh'' and ``Cyberpunk'' styles, confirm its broad applicability. More visual results are presented in the supplementary video.

\section{Conclusion}
In this paper, we present CustomTex, a novel framework for high-fidelity and instance-level controllable texturing of 3D indoor scenes. By providing a direct path from a set of reference images to a high-quality, unified texture map, CustomTex offers a more practical and user-friendly paradigm for 3D scene appearance editing. CustomTex has certain limitations we plan to address in future work. The training process to generate 4K-resolution textures is computationally intensive, typically taking several hours to complete. Furthermore, our method currently focuses on diffuse albedo texturing and does not generate other material maps, such as normal or roughness. 

{
    \small
    \bibliographystyle{ieeenat_fullname}
    \bibliography{reference}

@String(CVPR= {IEEE Conf. Comput. Vis. Pattern Recog.})

@String(ICCV= {Int. Conf. Comput. Vis.})

@String(ECCV= {Eur. Conf. Comput. Vis.})

@String(ICME = {Int. Conf. Multimedia and Expo})

@String(ICLR = {Int. Conf. Learn. Represent.})

@String(AAAI = {AAAI})

@String(CVPR  = {CVPR})

@String(ICCV  = {ICCV})

@String(ECCV  = {ECCV})

@String(ICME  =	{ICME})

@String(ICLR  = {ICLR})

@inproceedings{voxel1,
  author       = {Edward J. Smith and
                  David Meger},
  title        = {Improved Adversarial Systems for 3D Object Generation and Reconstruction},
  booktitle    = {Proc. of CORL},
  pages        = {87--96},
  year         = {2017}
}

@inproceedings{voxel2,
  author       = {Jianwen Xie and
                  Zilong Zheng and
                  Ruiqi Gao and
                  Wenguan Wang and
                  Song{-}Chun Zhu and
                  Ying Nian Wu},
  title        = {Learning Descriptor Networks for 3D Shape Synthesis and Analysis},
  booktitle    = {Proc. of CVPR},
  pages        = {8629--8638},
  year         = {2018}
}

@inproceedings{voxel3,
  author       = {Wenzheng Chen and
                  Huan Ling and
                  Jun Gao and
                  Edward J. Smith and
                  Jaakko Lehtinen and
                  Alec Jacobson and
                  Sanja Fidler},
  title        = {Learning to Predict 3D Objects with an Interpolation-based Differentiable
                  Renderer},
  booktitle    = {Proc. of NeurIPS},
  pages        = {9605--9616},
  year         = {2019}
}

@inproceedings{voxel4,
  author       = {Chieh Hubert Lin and
                  Hsin{-}Ying Lee and
                  Willi Menapace and
                  Menglei Chai and
                  Aliaksandr Siarohin and
                  Ming{-}Hsuan Yang and
                  Sergey Tulyakov},
  title        = {InfiniCity: Infinite-Scale City Synthesis},
  booktitle    = {Proc. of ICCV},
  pages        = {22751--22761},
  year         = {2023}
}

@inproceedings{voxel5,
  author       = {Aliaksandr Siarohin and
                  Willi Menapace and
                  Ivan Skorokhodov and
                  Kyle Olszewski and
                  Jian Ren and
                  Hsin{-}Ying Lee and
                  Menglei Chai and
                  Sergey Tulyakov},
  title        = {Unsupervised Volumetric Animation},
  booktitle    = {Proc. of CVPR},
  pages        = {458--469},
  year         = {2023}
}

@inproceedings{pointcloud1,
  author       = {Panos Achlioptas and
                  Olga Diamanti and
                  Ioannis Mitliagkas and
                  Leonidas J. Guibas},
  title        = {Learning Representations and Generative Models for 3D Point Clouds},
  booktitle    = {Proc. of ICLR},
  year         = {2018}
}

@inproceedings{pointcloud2,
  author       = {Linqi Zhou and
                  Yilun Du and
                  Jiajun Wu},
  title        = {3D Shape Generation and Completion through Point-Voxel Diffusion},
  booktitle    = {Proc. of ICCV},
  pages        = {5806--5815},
  year         = {2021}
}

@inproceedings{pointcloud3,
  author       = {Guandao Yang and
                  Xun Huang and
                  Zekun Hao and
                  Ming{-}Yu Liu and
                  Serge J. Belongie and
                  Bharath Hariharan},
  title        = {PointFlow: 3D Point Cloud Generation With Continuous Normalizing Flows},
  booktitle    = {Proc. of ICCV},
  pages        = {4540--4549},
  year         = {2019}
}

@article{mesh1,
  author       = {Lin Gao and
                  Jie Yang and
                  Tong Wu and
                  Yu{-}Jie Yuan and
                  Hongbo Fu and
                  Yu{-}Kun Lai and
                  Hao Zhang},
  title        = {{SDM-NET:} deep generative network for structured deformable mesh},
  journal      = {{ACM} Trans. Graph.},
  volume       = {38},
  pages        = {243:1--243:15},
  year         = {2019}
}

@article{mesh2,
  author       = {Jie Yang and
                  Kaichun Mo and
                  Yu{-}Kun Lai and
                  Leonidas J. Guibas and
                  Lin Gao},
  title        = {DSG-Net: Learning Disentangled Structure and Geometry for 3D Shape
                  Generation},
  journal      = {{ACM} Trans. Graph.},
  volume       = {42},
  pages        = {1:1--1:17},
  year         = {2023}
}

@inproceedings{mesh3,
  author       = {Dario Pavllo and
                  Graham Spinks and
                  Thomas Hofmann and
                  Marie{-}Francine Moens and
                  Aur{\'{e}}lien Lucchi},
  title        = {Convolutional Generation of Textured 3D Meshes},
  booktitle    = {Proc. of NeurIPS},
  year         = {2020}
}

@inproceedings{sdf1,
  author       = {Zhiqin Chen and
                  Hao Zhang},
  title        = {Learning Implicit Fields for Generative Shape Modeling},
  booktitle    = {Proc. of CVPR},
  pages        = {5939--5948},
  year         = {2019}
}

@article{sdf2,
  author       = {Amir Hertz and
                  Or Perel and
                  Raja Giryes and
                  Olga Sorkine{-}Hornung and
                  Daniel Cohen{-}Or},
  title        = {{SPAGHETTI:} editing implicit shapes through part aware generation},
  journal      = {{ACM} Trans. Graph.},
  volume       = {41},
  pages        = {106:1--106:20},
  year         = {2022}
}

@inproceedings{sdf3,
  author       = {Jeong Joon Park and
                  Peter R. Florence and
                  Julian Straub and
                  Richard A. Newcombe and
                  Steven Lovegrove},
  title        = {DeepSDF: Learning Continuous Signed Distance Functions for Shape Representation},
  booktitle    = {Proc. of CVPR},
  pages        = {165--174},
  year         = {2019}
}

@inproceedings{sdf4,
  author       = {Yen{-}Chi Cheng and
                  Hsin{-}Ying Lee and
                  Sergey Tulyakov and
                  Alexander G. Schwing and
                  Liangyan Gui},
  title        = {SDFusion: Multimodal 3D Shape Completion, Reconstruction, and Generation},
  booktitle    = {Proc. of CVPR},
  pages        = {4456--4465},
  year         = {2023}
}

@inproceedings{sdf5,
  author       = {Zezhou Cheng and
                  Menglei Chai and
                  Jian Ren and
                  Hsin{-}Ying Lee and
                  Kyle Olszewski and
                  Zeng Huang and
                  Subhransu Maji and
                  Sergey Tulyakov},
  title        = {Cross-modal 3D Shape Generation and Manipulation},
  booktitle    = {Proc. of ECCV},
  volume       = {13663},
  pages        = {303--321},
  year         = {2022}
}

@inproceedings{vlm1,
  author       = {Dave Zhenyu Chen and
                  Angel X. Chang and
                  Matthias Nie{\ss}ner},
  title        = {ScanRefer: 3D Object Localization in {RGB-D} Scans Using Natural Language},
  booktitle    = {Proc. of ECCV},
  pages        = {202--221},
  year         = {2020}
}

@inproceedings{vlm2,
  author       = {Dave Zhenyu Chen and
                  Ronghang Hu and
                  Xinlei Chen and
                  Matthias Nie{\ss}ner and
                  Angel X. Chang},
  title        = {UniT3D: {A} Unified Transformer for 3D Dense Captioning and Visual
                  Grounding},
  booktitle    = {Proc. of ICCV},
  pages        = {18063--18073},
  year         = {2023},
}

@inproceedings{vlm3,
  author       = {Dave Zhenyu Chen and
                  Qirui Wu and
                  Matthias Nie{\ss}ner and
                  Angel X. Chang},
  title        = {D\({}^{\mbox{3}}\)Net: {A} Unified Speaker-Listener Architecture for
                  3D Dense Captioning and Visual Grounding},
  booktitle    = {Proc. of ECCV},
  pages        = {487--505},
  year         = {2022}
}

@inproceedings{vlm4,
  author       = {Dave Zhenyu Chen and
                  Ali Gholami and
                  Matthias Nie{\ss}ner and
                  Angel X. Chang},
  title        = {Scan2Cap: Context-Aware Dense Captioning in {RGB-D} Scans},
  booktitle    = {Proc. of CVPR},
  pages        = {3193--3203},
  year         = {2021}
}

@inproceedings{vlm5,
  author       = {Ronghang Hu and
                  Amanpreet Singh},
  title        = {UniT: Multimodal Multitask Learning with a Unified Transformer},
  booktitle    = {Proc. of ICCV},
  pages        = {1419--1429},
  publisher    = {{IEEE}},
  year         = {2021}
}

@inproceedings{2DI1,
  author       = {Rameen Abdal and
                  Hsin{-}Ying Lee and
                  Peihao Zhu and
                  Menglei Chai and
                  Aliaksandr Siarohin and
                  Peter Wonka and
                  Sergey Tulyakov},
  title        = {3DAvatarGAN: Bridging Domains for Personalized Editable Avatars},
  booktitle    = {Proc. of CVPR},
  pages        = {4552--4562},
  year         = {2023}
}

@inproceedings{2DI2,
  author       = {Eric R. Chan and
                  Marco Monteiro and
                  Petr Kellnhofer and
                  Jiajun Wu and
                  Gordon Wetzstein},
  title        = {Pi-GAN: Periodic Implicit Generative Adversarial Networks for 3D-Aware
                  Image Synthesis},
  booktitle    = {Proc. of CVPR},
  pages        = {5799--5809},
  year         = {2021}
}

@inproceedings{2DI3,
  author       = {Eric R. Chan and
                  Connor Z. Lin and
                  Matthew A. Chan and
                  Koki Nagano and
                  Boxiao Pan and
                  Shalini De Mello and
                  Orazio Gallo and
                  Leonidas J. Guibas and
                  Jonathan Tremblay and
                  Sameh Khamis and
                  Tero Karras and
                  Gordon Wetzstein},
  title        = {Efficient Geometry-aware 3D Generative Adversarial Networks},
  booktitle    = {Proc. of CVPR},
  pages        = {16102--16112},
  year         = {2022}
}

@inproceedings{2DI4,
  author       = {Norman M{\"{u}}ller and
                  Yawar Siddiqui and
                  Lorenzo Porzi and
                  Samuel Rota Bul{\`{o}} and
                  Peter Kontschieder and
                  Matthias Nie{\ss}ner},
  title        = {DiffRF: Rendering-Guided 3D Radiance Field Diffusion},
  booktitle    = {Proc. of CVPR},
  pages        = {4328--4338},
  year         = {2023}
}

@inproceedings{2DI6,
  author       = {Jun Gao and
                  Tianchang Shen and
                  Zian Wang and
                  Wenzheng Chen and
                  Kangxue Yin and
                  Daiqing Li and
                  Or Litany and
                  Zan Gojcic and
                  Sanja Fidler},
  editor       = {Sanmi Koyejo and
                  S. Mohamed and
                  A. Agarwal and
                  Danielle Belgrave and
                  K. Cho and
                  A. Oh},
  title        = {{GET3D:} {A} Generative Model of High Quality 3D Textured Shapes Learned
                  from Images},
  booktitle    = {Proc. of NeurIPS},
  year         = {2022}
}

@article{geometry1,
  title={A statistical model for synthesis of detailed facial geometry},
  author={Golovinskiy, Aleksey and Matusik, Wojciech and Pfister, Hanspeter and Rusinkiewicz, Szymon and Funkhouser, Thomas},
  journal={{ACM} Trans. Graph.},
  volume={25},
  pages={1025--1034},
  year={2006}
}

@article{geometry2,
  title={A hierarchical extension to 3D non-parametric surface relief completion},
  author={Breckon, Toby P and Fisher, Robert B},
  journal={Pattern Recognit.},
  volume={45},
  pages={172--185},
  year={2012}
}

@article{geometry3,
  author       = {Amir Hertz and
                  Rana Hanocka and
                  Raja Giryes and
                  Daniel Cohen{-}Or},
  title        = {Deep geometric texture synthesis},
  journal      = {{ACM} Trans. Graph.},
  volume       = {39},
  pages        = {108},
  year         = {2020}
}

@article{color1,
  author       = {Jianye Lu and
                  Athinodoros S. Georghiades and
                  Andreas Glaser and
                  Hongzhi Wu and
                  Li{-}Yi Wei and
                  Baining Guo and
                  Julie Dorsey and
                  Holly E. Rushmeier},
  title        = {Context-aware textures},
  journal      = {{ACM} Trans. Graph.},
  volume       = {26},
  pages        = {3},
  year         = {2007}
}

@inproceedings{2dt1,
  title={Multiresolution sampling procedure for analysis and synthesis of texture images},
  author={De Bonet, Jeremy S},
  booktitle={Proc. of SIGGRAPH},
  pages={361--368},
  year={1997}
}

@inproceedings{2dt2,
  title={Texture synthesis by non-parametric sampling},
  author={Efros, Alexei A and Leung, Thomas K},
  booktitle={Proc. of ICCV},
  pages={1033--1038},
  year={1999}
}

@inproceedings{uv,
  title={Auv-net: Learning aligned uv maps for texture transfer and synthesis},
  author={Chen, Zhiqin and Yin, Kangxue and Fidler, Sanja},
  booktitle={Proc. of CVPR},
  pages={1455--1464},
  year={2022}
}

@inproceedings{c1,
  title={Texturify: Generating textures on 3d shape surfaces},
  author={Siddiqui, Yawar and Thies, Justus and Ma, Fangchang and Shan, Qi and Nie{\ss}ner, Matthias and Dai, Angela},
  booktitle={Proc. of ECCV},
  pages={72--88},
  year={2022}
}

@inproceedings{c2,
  title={Mesh2tex: Generating mesh textures from image queries},
  author={Bokhovkin, Alexey and Tulsiani, Shubham and Dai, Angela},
  booktitle={Proc. of ICCV},
  pages={8884--8894},
  year={2023}
}

@inproceedings{stylegan,
  title={Analyzing and improving the image quality of stylegan},
  author={Karras, Tero and Laine, Samuli and Aittala, Miika and Hellsten, Janne and Lehtinen, Jaakko and Aila, Timo},
  booktitle={Proc. of CVPR},
  pages={8107--8116},
  year={2020}
}

@article{transformer1,
  title={Triposr: Fast 3d object reconstruction from a single image},
  author={Tochilkin, Dmitry and Pankratz, David and Liu, Zexiang and Huang, Zixuan and Letts, Adam and Li, Yangguang and Liang, Ding and Laforte, Christian and Jampani, Varun and Cao, Yan-Pei},
  journal={arXiv preprint arXiv:2403.02151},
  year={2024}
}

@inproceedings{transformer2,
  title={Grm: Large gaussian reconstruction model for efficient 3d reconstruction and generation},
  author={Xu, Yinghao and Shi, Zifan and Yifan, Wang and Chen, Hansheng and Yang, Ceyuan and Peng, Sida and Shen, Yujun and Wetzstein, Gordon},
  booktitle={Proc. of ECCV},
  pages={1--20},
  year={2024}
}

@inproceedings{diffusion6,
  author       = {Jonathan Ho and
                  Ajay Jain and
                  Pieter Abbeel},
  editor       = {Hugo Larochelle and
                  Marc'Aurelio Ranzato and
                  Raia Hadsell and
                  Maria{-}Florina Balcan and
                  Hsuan{-}Tien Lin},
  title        = {Denoising Diffusion Probabilistic Models},
  booktitle    = {Proc. of NeurIPS},
  year         = {2020},
}

@inproceedings{object1,
  title={AE-NeRF: Augmenting Event-Based Neural Radiance Fields for Non-ideal Conditions and Larger Scenes},
  author={Feng, Chaoran and Yu, Wangbo and Cheng, Xinhua and Tang, Zhenyu and Zhang, Junwu and Yuan, Li and Tian, Yonghong},
  booktitle={Proc. of AAAI},
  pages={2924--2932},
  year={2025}
}

@inproceedings{object2,
  author       = {Yu, Wangbo and Feng, Chaoran and Li, Jianing and Tang, Jiye and Yang, Jiashu and Tang, Zhenyu and Cao, Meng and Jia, Xu and Yang, Yuchao and Yuan, Li and others},
  title        = {EvaGaussians: Event Stream Assisted Gaussian Splatting from Blurry Images},
  booktitle={Proc. of ICCV},
  pages={24780--24790},
  year={2025}
}

@inproceedings{object3,
  author       = {Zhengyi Wang and
                  Cheng Lu and
                  Yikai Wang and
                  Fan Bao and
                  Chongxuan Li and
                  Hang Su and
                  Jun Zhu},
  title        = {ProlificDreamer: High-Fidelity and Diverse Text-to-3D Generation with
                  Variational Score Distillation},
  booktitle    = {Proc. of NeurIPS},
  year         = {2023},
}

@article{scene1,
  title={Text2nerf: Text-driven 3d scene generation with neural radiance fields},
  author={Zhang, Jingbo and Li, Xiaoyu and Wan, Ziyu and Wang, Can and Liao, Jing},
  journal={IEEE Trans. Vis. Comput. Graph.},
  volume={30},
  number={12},
  pages={7749--7762},
  year={2024}
}

@inproceedings{scene2,
  title={Roompainter: View-integrated diffusion for consistent indoor scene texturing},
  author={Huang, Zhipeng and Yu, Wangbo and Cheng, Xinhua and Zhao, ChengShu and Ge, Yunyang and Guo, Mingyi and Yuan, Li and Tian, Yonghong},
  booktitle={Proc. of CVPR},
  pages={574--584},
  year={2025}
}

@inproceedings{Text2Tex,
  title={Text2tex: Text-driven texture synthesis via diffusion models},
  author={Chen, Dave Zhenyu and Siddiqui, Yawar and Lee, Hsin-Ying and Tulyakov, Sergey and Nie{\ss}ner, Matthias},
  booktitle={Proc. of ICCV},
  pages={18512--18522},
  year={2023}
}

@inproceedings{TEXTure,
  title={Texture: Text-guided texturing of 3d shapes},
  author={Richardson, Elad and Metzer, Gal and Alaluf, Yuval and Giryes, Raja and Cohen-Or, Daniel},
  booktitle={Proc. of ACM SIGGRAPH},
  pages={54:1--54:11},
  year={2023}
}

@inproceedings{TexFusion,
  title={Texfusion: Synthesizing 3d textures with text-guided image diffusion models},
  author={Cao, Tianshi and Kreis, Karsten and Fidler, Sanja and Sharp, Nicholas and Yin, Kangxue},
  booktitle={Proc. of ICCV},
  pages={4146--4158},
  year={2023}
}

@inproceedings{depthad1,
  title={T2i-adapter: Learning adapters to dig out more controllable ability for text-to-image diffusion models},
  author={Mou, Chong and Wang, Xintao and Xie, Liangbin and Wu, Yanze and Zhang, Jian and Qi, Zhongang and Shan, Ying},
  booktitle={Proc. of AAAI},
  number={5},
  pages={4296--4304},
  year={2024}
}

@inproceedings{depthad2,
  title={Adding conditional control to text-to-image diffusion models},
  author={Zhang, Lvmin and Rao, Anyi and Agrawala, Maneesh},
  booktitle={Proc. of ICCV},
  pages={3813--3824},
  year={2023}
}

@inproceedings{genesistex,
  title={Genesistex: adapting image denoising diffusion to texture space},
  author={Gao, Chenjian and Jiang, Boyan and Li, Xinghui and Zhang, Yingpeng and Yu, Qian},
  booktitle={Proc. of CVPR},
  pages={4620--4629},
  year={2024}
}

@inproceedings{genesistex2,
  title={Genesistex2: Stable, consistent and high-quality text-to-texture generation},
  author={Lu, Jiawei and Zhang, Yingpeng and Zhao, Zengjun and Wang, He and Zhou, Kun and Shao, Tianjia},
  booktitle={Proc. of AAAI},
  volume={39},
  pages={5820--5828},
  year={2025}
}

@inproceedings{dreamfusion,
  author       = {Ben Poole and
                  Ajay Jain and
                  Jonathan T. Barron and
                  Ben Mildenhall},
  title        = {DreamFusion: Text-to-3D using 2D Diffusion},
  booktitle    = {Proc. of ICLR},
  year         = {2023}
}

@inproceedings{latentnerf,
  title={Latent-nerf for shape-guided generation of 3d shapes and textures},
  author={Metzer, Gal and Richardson, Elad and Patashnik, Or and Giryes, Raja and Cohen-Or, Daniel},
  booktitle={Proc. of CVPR},
  pages={12663--12673},
  year={2023}
}

@inproceedings{sds1,
  title={Fantasia3d: Disentangling geometry and appearance for high-quality text-to-3d content creation},
  author={Chen, Rui and Chen, Yongwei and Jiao, Ningxin and Jia, Kui},
  booktitle={Proc. of ICCV},
  pages={22189--22199},
  year={2023}
}

@inproceedings{sds2,
  title={Paint-it: Text-to-texture synthesis via deep convolutional texture map optimization and physically-based rendering},
  author={Youwang, Kim and Oh, Tae-Hyun and Pons-Moll, Gerard},
  booktitle={Proc. of CVPR},
  pages={4347--4356},
  year={2024}
}

@inproceedings{sds3,
  title={Magic3d: High-resolution text-to-3d content creation},
  author={Lin, Chen-Hsuan and Gao, Jun and Tang, Luming and Takikawa, Towaki and Zeng, Xiaohui and Huang, Xun and Kreis, Karsten and Fidler, Sanja and Liu, Ming-Yu and Lin, Tsung-Yi},
  booktitle={Proc. of CVPR},
  pages={300--309},
  year={2023}
}

@inproceedings{SceneTex2024,
  author       = {Dave Zhenyu Chen and
                  Haoxuan Li and
                  Hsin{-}Ying Lee and
                  Sergey Tulyakov and
                  Matthias Nie{\ss}ner},
  title        = {SceneTex: High-Quality Texture Synthesis for Indoor Scenes via Diffusion
                  Priors},
  booktitle    = {Proc. of CVPR},
  pages        = {21081--21091},
  year         = {2024},
}

@inproceedings{NeRF2020,
  author       = {Ben Mildenhall and
                  Pratul P. Srinivasan and
                  Matthew Tancik and
                  Jonathan T. Barron and
                  Ravi Ramamoorthi and
                  Ren Ng},
  editor       = {Andrea Vedaldi and
                  Horst Bischof and
                  Thomas Brox and
                  Jan{-}Michael Frahm},
  title        = {NeRF: Representing Scenes as Neural Radiance Fields for View Synthesis},
  booktitle    = {Proc. of ECCV},
  volume       = {12346},
  pages        = {405--421},
  year         = {2020},
}

@article{3DGSTOG2023,
  author       = {Bernhard Kerbl and
                  Georgios Kopanas and
                  Thomas Leimk{\"{u}}hler and
                  George Drettakis},
  title        = {3D Gaussian Splatting for Real-Time Radiance Field Rendering},
  journal      = {{ACM} Trans. Graph.},
  volume       = {42},
  number       = {4},
  pages        = {139:1--139:14},
  year         = {2023}
}

@inproceedings{StableDiffusion2022,
  author       = {Robin Rombach and
                  Andreas Blattmann and
                  Dominik Lorenz and
                  Patrick Esser and
                  Bj{\"{o}}rn Ommer},
  title        = {High-Resolution Image Synthesis with Latent Diffusion Models},
  booktitle    = {Proc. of CVPR},
  pages        = {10674--10685},
  year         = {2022},
}

@inproceedings{FlexiTex2025,
  author       = {Dadong Jiang and
                  Xianghui Yang and
                  Zibo Zhao and
                  Sheng Zhang and
                  Jiaao Yu and
                  Zeqiang Lai and
                  Shaoxiong Yang and
                  Chunchao Guo and
                  Xiaobo Zhou and
                  Zhihui Ke},
  editor       = {Toby Walsh and
                  Julie Shah and
                  Zico Kolter},
  title        = {FlexiTex: Enhancing Texture Generation via Visual Guidance},
  booktitle    = {Proc. of AAAI},
  pages        = {3967--3975},
  year         = {2025},
}

@inproceedings{InstanceTex2024,
  author       = {Mingxin Yang and
                  Jianwei Guo and
                  Yuzhi Chen and
                  Lan Chen and
                  Pu Li and
                  Zhanglin Cheng and
                  Xiaopeng Zhang and
                  Hui Huang},
  editor       = {Takeo Igarashi and
                  Ariel Shamir and
                  Hao (Richard) Zhang},
  title        = {InstanceTex: Instance-level Controllable Texture Synthesis for 3D
                  Scenes via Diffusion Priors},
  booktitle    = {Proc. of SIGGRAPH Asia},
  pages        = {59:1--59:11},
  year         = {2024},
}

@inproceedings{PiSA-SR2025,
  author       = {Lingchen Sun and
                  Rongyuan Wu and
                  Zhiyuan Ma and
                  Shuaizheng Liu and
                  Qiaosi Yi and
                  Lei Zhang},
  title        = {Pixel-level and Semantic-level Adjustable Super-resolution: {A} Dual-LoRA
                  Approach},
  booktitle    = {Proc. of CVPR},
  pages        = {2333--2343},
  year         = {2025},
}

@inproceedings{Paint3D,
  title={Paint3d: Paint anything 3d with lighting-less texture diffusion models},
  author={Zeng, Xianfang and Chen, Xin and Qi, Zhongqi and Liu, Wen and Zhao, Zibo and Wang, Zhibin and Fu, Bin and Liu, Yong and Yu, Gang},
  booktitle={Proc. of CVPR},
  pages={4252--4262},
  year={2024}
}

@article{HY3D-2.1,
    title={Hunyuan3D 2.1: From Images to High-Fidelity 3D Assets with Production-Ready PBR Material},
    author={Tencent Hunyuan3D Team},
    journal={arXiv preprint arXiv:2506.15442},
    year={2025}
}

@inproceedings{SyncMVD,
    author = {Liu, Yuxin and Xie, Minshan and Liu, Hanyuan and Wong, Tien-Tsin},
    title = {Text-Guided Texturing by Synchronized Multi-View Diffusion},
    booktitle = {Proc. of SIGGRAPH Asia},
    pages = {60:1--60:11},
    year = {2024},
}

@article{IPAdapter,
  title={IP-Adapter: Text Compatible Image Prompt Adapter for Text-to-Image Diffusion Models},
  author={Hu Ye and Jun Zhang and Siyi Liu and Xiao Han and Wei Yang},
  journal={arXiv preprint arXiv:2308.06721},
  year={2023}
}

@inproceedings{CLIP,
  title={Learning transferable visual models from natural language supervision},
  author={Radford, Alec and Kim, Jong Wook and Hallacy, Chris and Ramesh, Aditya and Goh, Gabriel and Agarwal, Sandhini and Sastry, Girish and Askell, Amanda and Mishkin, Pamela and Clark, Jack and others},
  booktitle={Proc. of ICML},
  pages={8748--8763},
  year={2021},
}

@inproceedings{FID,
  title={On aliased resizing and surprising subtleties in gan evaluation},
  author={Parmar, Gaurav and Zhang, Richard and Zhu, Jun-Yan},
  booktitle={Proc. of CVPR},
  pages={11400--11410},
  year={2022}
}

@inproceedings{Qalign,
  title={Q-align: Teaching lmms for visual scoring via discrete text-defined levels},
  author={Wu, Haoning and Zhang, Zicheng and Zhang, Weixia and Chen, Chaofeng and Liao, Liang and Li, Chunyi and Gao, Yixuan and Wang, Annan and Zhang, Erli and Sun, Wenxiu and others},
  booktitle={Proc. of ICML},
  pages={54015--54029},
  year={2024}
}

@article{iNGP,
  title={Instant neural graphics primitives with a multiresolution hash encoding},
  author={M{\"u}ller, Thomas and Evans, Alex and Schied, Christoph and Keller, Alexander},
  journal={{ACM} Trans. Graph.},
  volume={41},
  number={4},
  pages={102:1--102:15},
  year={2022},
}

@inproceedings{3D-FRONT2021,
  author       = {Huan Fu and
                  Bowen Cai and
                  Lin Gao and
                  Lingxiao Zhang and
                  Jiaming Wang and
                  Cao Li and
                  Qixun Zeng and
                  Chengyue Sun and
                  Rongfei Jia and
                  Binqiang Zhao and
                  Hao Zhang},
  title        = {3D-FRONT: 3D Furnished Rooms with layOuts and semaNTics},
  booktitle    = {Proc. of ICCV},
  pages        = {10913--10922},
  year         = {2021},
}

@article{FaceRefiner2024,
  author       = {Chengyang Li and
                  Baoping Cheng and
                  Yao Cheng and
                  Haocheng Zhang and
                  Renshuai Liu and
                  Yinglin Zheng and
                  Jing Liao and
                  Xuan Cheng},
  title        = {FaceRefiner: High-Fidelity Facial Texture Refinement With Differentiable
                  Rendering-Based Style Transfer},
  journal      = {{IEEE} Trans. Multim.},
  volume       = {26},
  pages        = {7225--7236},
  year         = {2024}
}

@article{DCT2024,
  author       = {Renshuai Liu and
                  Yao Cheng and
                  Sifei Huang and
                  Chengyang Li and
                  Xuan Cheng},
  title        = {Transformer-Based High-Fidelity Facial Displacement Completion for
                  Detailed 3D Face Reconstruction},
  journal      = {{IEEE} Trans. Multim.},
  volume       = {26},
  pages        = {799--810},
  year         = {2024}
}

@inproceedings{DiffSFSR2024,
  author       = {Renshuai Liu and
                  Bowen Ma and
                  Wei Zhang and
                  Zhipeng Hu and
                  Changjie Fan and
                  Tangjie Lv and
                  Yu Ding and
                  Xuan Cheng},
  title        = {Towards a Simultaneous and Granular Identity-Expression Control in
                  Personalized Face Generation},
  booktitle    = {Proc. of CVPR},
  pages        = {2114--2123},
  year         = {2024},
}

@article{SizeScene2025,
  author       = {Yao Cheng and
                  Weilin Chen and
                  Yizhe Gu and
                  Yue Sun and
                  You Zhai and
                  Xuan Cheng and
                  Juncong Lin},
  title        = {Size-aware indoor scene retargeting with generalized summarization},
  journal      = {Comput. Graph.},
  volume       = {132},
  pages        = {104315},
  year         = {2025},
}

@article{Learn2Talk2025,
  author       = {Yixiang Zhuang and
                  Baoping Cheng and
                  Yao Cheng and
                  Yuntao Jin and
                  Renshuai Liu and
                  Chengyang Li and
                  Xuan Cheng and
                  Jing Liao and
                  Juncong Lin},
  title        = {Learn2Talk: 3D Talking Face Learns From 2D Talking Face},
  journal      = {{IEEE} Trans. Vis. Comput. Graph.},
  volume       = {31},
  number       = {9},
  pages        = {5829--5841},
  year         = {2025}
}

@inproceedings{DNPM,
  author       = {Haitao Cao and
                  Baoping Cheng and
                  Qiran Pu and
                  Haocheng Zhang and
                  Bin Luo and
                  Yixiang Zhuang and
                  Juncong Lin and
                  Liyan Chen and
                  Xuan Cheng},
  title        = {{DNPM:} {A} Neural Parametric Model for the Synthesis of Facial Geometric
                  Details},
  booktitle    = {Proc. of ICME},
  pages        = {1--6},
  year         = {2024},
}

@article{PointsL0,
  author       = {Xuan Cheng and
                  Ming Zeng and
                  Jinpeng Lin and
                  Zizhao Wu and
                  Xinguo Liu},
  title        = {Efficient L0 resampling of point sets},
  journal      = {Comput. Aided Geom. Des.},
  volume       = {75},
  year         = {2019},

}

@inproceedings{EMEF2023,
  author       = {Renshuai Liu and
                  Chengyang Li and
                  Haitao Cao and
                  Yinglin Zheng and
                  Ming Zeng and
                  Xuan Cheng},
  editor       = {Brian Williams and
                  Yiling Chen and
                  Jennifer Neville},
  title        = {{EMEF:} Ensemble Multi-Exposure Image Fusion},
  booktitle    = {Proc. of AAAI},
  pages        = {1710--1718},
  year         = {2023}
}

@inproceedings{zhang2025storyweaver,
  title={Storyweaver: A unified world model for knowledge-enhanced story character customization},
  author={Zhang, Jinlu and Tang, Jiji and Zhang, Rongsheng and Lv, Tangjie and Sun, Xiaoshuai},
  booktitle={Proc. of AAAI},
  volume={39},
  number={9},
  pages={9951--9959},
  year={2025}
}

@InProceedings{zhang2024fast,
  title = 	 {Fast Text-to-3{D}-Aware Face Generation and Manipulation via Direct Cross-modal Mapping and Geometric Regularization},
  author =       {Zhang, Jinlu and Zhou, Yiyi and Zheng, Qiancheng and Du, Xiaoxiong and Luo, Gen and Peng, Jun and Sun, Xiaoshuai and Ji, Rongrong},
  booktitle = 	 {Proc. of ICML},
  pages = 	 {60605--60625},
  year = 	 {2024},
  volume = 	 {235}
}
}

\end{document}